\documentclass[journal]{IEEEtran}

\usepackage{graphicx}
\usepackage{stfloats}

\pdfoutput=1

\usepackage{mathtools}
\usepackage{amsmath}
\usepackage{amsthm}
\usepackage{amssymb}
\usepackage{amsfonts}
\usepackage{amsopn}
\usepackage{graphicx} 
\usepackage{textcomp}
\usepackage{xfrac}
\usepackage{bbm}
\usepackage{overpic}
\usepackage{subfig}
\usepackage{wrapfig}
\usepackage[ruled,vlined,linesnumbered]{algorithm2e}
\usepackage{multirow}

\usepackage{color}
\definecolor{green}{rgb}{0, 0.5, 0}
\definecolor{orange}{rgb}{0.8, 0.6, 0.2}
\definecolor{red}{rgb}{1.0, 0.0, 0.0}
\definecolor{teal}{rgb}{0.0, 0.4, 0.4}
\definecolor{purple}{rgb}{0.65,0,0.65}
\definecolor{saffron}{rgb}{0.95,0.75,0.2}
\definecolor{turquoise}{rgb}{0.0,0.5,0.5}
\definecolor{brown}{rgb}{0.5, 0.16, 0.16}

\usepackage{overpic}
\usepackage{currfile} 

\newcommand{\myfigurename}{\put(-3,0){\vertical{\todo{\currfiledir}}}}
\renewcommand{\myfigurename}{}

\newcommand{\pd}[1]{{\color{green}#1}}

\definecolor{lightgray}{rgb}{0.6, 0.6, 0.6}

\newcommand{\Fig}[1]{Fig.~\ref{fig:#1}}
\newcommand{\Eq}[1]{Eq.~(\ref{eq:#1})}

\newcommand{\Sec}[1]{Section~\ref{sec:#1}}
\newcommand{\Algo}[1]{Algorithm~\ref{algo:#1}}
\newcommand{\Table}[1]{Table~\ref{tab:#1}}

\usepackage[normalem]{ulem}

\renewcommand{\paragraph}[1]{\textbf{#1.}}

\newcommand{\hidecomment}[1]{}

\DeclareMathOperator*{\argmax}{arg\,max}

\usepackage{xspace}

\newcommand{\obf}{observability field\xspace}

\begin{document}

\title{Autonomous Outdoor Scanning via Online Topological and Geometric Path Optimization}

\author{Pengdi~Huang, Liqiang~Lin, Kai~Xu, and Hui~Huang

	\thanks{Pengdi Huang, Liqiang Lin, and Hui Huang are with Visual Computing Research Center, Shenzhen University, Shenzhen 518060, China (email: alualu628628@gmail.com; liniquie@gmail.com; hhzhiyan@gmail.com)}
	\thanks{Kai Xu is with School of Computer Science, National University of Defense Technology, Changsha 410000, China (email: kevin.kai.xu@gmail.com)}
    \thanks{$^\dagger$Corresponding author: Hui Huang}
}

\maketitle

\begin{abstract}
Autonomous 3D acquisition of outdoor environments poses special challenges. Different from indoor scenes, where the room space is delineated by clear boundaries and separations (e.g., walls and furniture), an outdoor environment is spacious and unbounded (thinking of a campus). Therefore, unlike for indoor scenes where the scanning effort is mainly devoted to the discovery of boundary surfaces, scanning an open and unbounded area requires actively delimiting the extent of scanning region and dynamically planning a traverse path within that region. Thus, for outdoor scenes, we formulate the planning of an energy-efficient autonomous scanning through a discrete-continuous optimization of robot scanning paths. The discrete optimization computes a topological map, through solving an online traveling sales problem (Online TSP), which determines the scanning goals and paths on-the-fly. The dynamic goals are determined as a collection of visit sites with high reward of visibility-to-unknown. A visit graph is constructed via connecting the visit sites with edges weighted by traversing cost. This topological map evolves as the robot scans via deleting outdated sites that are either visited or become rewardless and inserting newly discovered ones. The continuous part optimizes the traverse paths geometrically between two neighboring visit sites via maximizing the information gain of scanning along the paths. The discrete and continuous processes alternate until the traverse cost of the current graph exceeds the remaining energy capacity of the robot. Our approach is evaluated with both synthetic and field tests, demonstrating its effectiveness and advantages over alternatives. The project is at http://vcc.szu.edu.cn/research/2020/Husky, and the codes are available at https://github.com/alualu628628/Autonomous-Outdoor-Scanning-via-Online-Topological-and-Geometric-Path-Optimization.
\end{abstract}


\IEEEpeerreviewmaketitle

\captionsetup[figure]{labelfont={bf},name={Fig.},labelsep=period}

\section{Introduction}
\label{sec:intro}

\IEEEPARstart{W}{ith} an increasing demand for digitized large-scale 3D scenes,
considerable research effort is being devoted to improving the scalability
and accuracy of 3D acquisition of scenes. For human-operated scanning, however, scalability and accuracy are inherently competing goals.
High quality acquisition requires smooth scanning trajectories and slow scanning motion, which make manual scanning rather laborious and time-consuming especially for large-scale scenes.
Moreover, it is challenging for a non-skilled user to find smooth trajectories ensuring high quality scanning.

Recently, autonomous scanning has gained increasing attention in the graphics field.
In this approach, a mobile robot or a quadrotor is driven to autonomously explore and/or reconstruct an environment with higher scalability and accuracy~\cite{Charrow2015,Song2015,Xu15,xu2017autonomous,liu2018,dong2019multirobot,hepp2018plan3d}.
The objective is that a robot leverages the progressively acquired scene geometry in the planning of movement paths and scanning trajectories for explorative acquisition.
The joint optimization of exploration and scanning trajectories provides highly effective 3D acquisition with high levels of scan coverage and quality. Nevertheless, existing systems have so far mainly focused on \emph{indoor} environments.

\begin{figure*}[t]
\begin{overpic}[width=0.99\textwidth,tics=5]{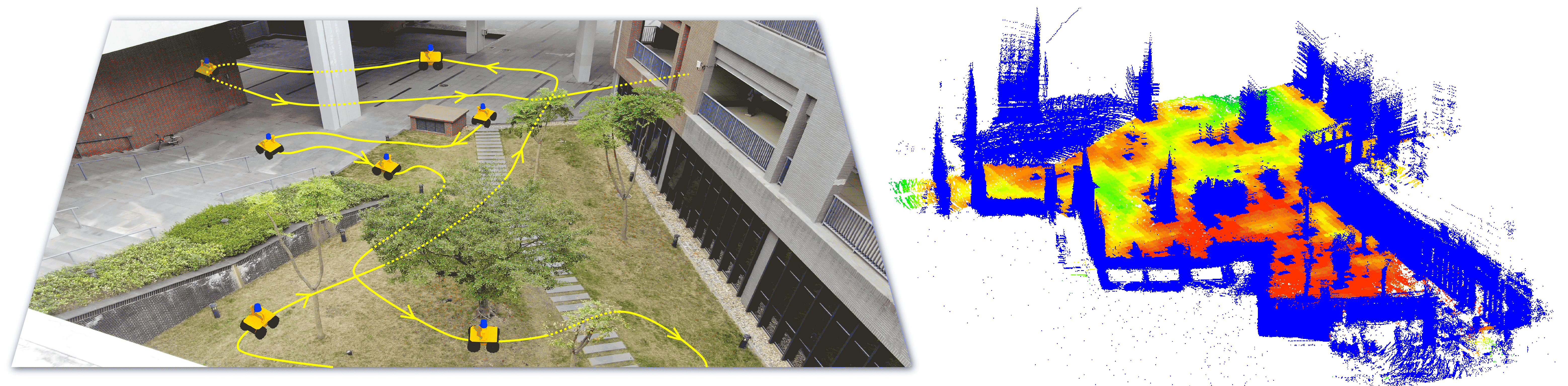}
	\put(2,2){\small (a)}
	\put(62,2){\small (b)}
\end{overpic}
\caption{
A mobile robot is autonomously exploring and scanning an unseen outdoor scene. The path (yellow curve in (a)) is planned and optimized on-the-fly, while the scans are incrementally aggregated. The final point cloud in (b) plots the scanning confidence (red is confident) over the ground.}
\label{fig:teaser}
\end{figure*}

Autonomous-scanning of \emph{outdoor} environments poses new and special challenges.
Different from indoor scenes, where the room space is delineated by clear boundaries and separations (e.g., walls and furniture) and the ground is usually flat, an outdoor environment is more spacious and usually boundless and the ground can be an arbitrary surface. One typical example is a spacious campus (see Fig.~\ref{fig:teaser}).
Therefore, unlike for indoor scenes, where the scanning effort is mainly devoted to the discovery of boundary surfaces, scanning an open and unbounded area requires actively delimiting the extent of scanning region and dynamically planning a traverse path within the already scanned open region.

In this work, we approach the planning of an auto-scanning course for efficient acquisition of outdoor scenes. Our key motivation is an on-the-fly planning of scanning path in an unknown environment. We formulate the dynamic planning as a discrete-continuous optimization of robot scanning paths. The discrete optimization computes a \emph{topological} guidance map and solves an Online Traveling Salesman Problem (Online TSP) for dynamically determined scanning goals and traverse paths.
The dynamic goals are defined on-the-fly, as a collection of visit sites, with high reward for sites which have more visibility of unknown regions. A visit-graph is constructed by connecting the visit sites with edges weighted by their traversing cost.
This topological map evolves as the robot scans via deleting outdated sites, which are either already visited or become reward-less and inserting newly discovered ones.
The continuous part, on the other hand, optimizes \emph{geometrically} the traverse paths between two sites visited consecutively along the TSP path, by maximizing the information gain of scanning along it.
The discrete and continuous processes alternate and refine each other until the traverse cost of the current graph exceeds the remaining energy capacity of the robot. Fig.~\ref{fig:teaser} visualizes the guidance map and traverse paths.


We develop an auto-scanning system for outdoor environments using the Husky robot mounted with a LiDAR scanner. Note, however, our method is not confined with a specific type of robot or scanner.
Our method is evaluated both with virtual run on synthetic 3D scenes and with field tests over real-world environments, all demonstrating promising results in both quantitative and qualitative evaluation.
We are able to show that our method achieves significantly higher acquisition quality than baseline or alternative methods.
In addition, we show that our method is extremely efficient and robust against robot initialization and scene complexity, showing its practical usability and scalability in scanning large outdoor environments.
To sum up, our contributions include:
\begin{itemize}
	\item A new formulation of the on-the-fly planning of quality scanning of outdoor scenes as the joint online optimization of the topology and geometry of robot paths.
	\item An online discrete-continuous planning, which encompasses a couple of novel technical components, such as dynamic path planning formulated as an online traveling salesman problem, unknown estimation based on point cloud visibility, and scan quality measurement based on fractal dimension. To the best of our knowledge, our work is the first that adapts and unifies these techniques into auto-scanning.
	\item An end-to-end scanning system for quality and efficient scanning of large, unknown outdoor scenes, as well as a benchmark for this task, which will all be released.
\end{itemize}

\section{Related work}
\label{sec:related}

\paragraph{Scene scanning and reconstruction}
For indoor scenes, we have seen significant advances in both online and offline RGB-D reconstruction methods,
with the introduction of the commodity depth camera.
KinectFusion~\cite{Newcombe2011,Izadi2011} is one of the first to realize a real-time volumetric fusion framework~\cite{Curless96}.
To handle larger environments, spatial hierarchies~\cite{chen2013scalable}, hashing schemes~\cite{Niessner2013,Kahler2015}, and rolling volumes~\cite{Whelan2012} have been proposed.
For scanning large scenes, global camera pose optimization is commonly used in offline approaches~\cite{choi2015}.
Outdoor scanning is typically performed with LiDAR scanners and the reconstruction usually conducted offline~\cite{musialski2013survey}.

\paragraph{Autonomous reconstruction with robots}
Recent years have witnessed fast development on autonomous scanning for scene reconstruction in the graphics community, where robots are planned and driven to scan the indoor environment.
In contrast to the traditional works on robot mapping~\cite{leonard1992dynamic,allen2001avenue,surmann2003autonomous}, the main focus of these works is on the final \emph{dense} surface reconstruction quality, rather than only creating a \emph{sparse} map.
The first methods started to look at a single object~\cite{Krainin2011,Kriegel2012,Wu2014}, which was then subsequently expanded to larger environments~\cite{Charrow2015}.
Very recently, we have seen results using time-varying tensor field optimization that achieve good reconstruction quality for room-size indoor environments~\cite{xu2017autonomous}.

In parallel to autonomous indoor reconstruction, there has been extensive research on outdoor 3D reconstruction, such as city-scale reconstruction, using laser scanners mounted on automobiles or quadrotors. However, these methods are typically not interactive and rely on off-line scan planning and/or scan registration for reconstruction~\cite{yu2015semantic,hepp2018plan3d,roberts2017submodular,musialski2013survey}.
Our method achieves online planning through a joint topological and geometric path optimization.
The work of~\cite{krusi2017driving} is the closest related to ours, where they optimize 6D trajectories compliant with curvature and continuity constraints of terrain directly over point cloud maps.
Their motion planning, however, mainly considers local traversibility, based on local planar patch fitting.
Our method, on the other hand, achieves both local and global planning by the discrete-continuous path optimization.

\paragraph{Next best view (NBV) planning}
NBV selection and camera trajectory optimization are the core problems for robot-operated autonomous scanning~\cite{chen2011active,ramanagopal2017motion}.
Due to the explorative nature of auto-scanning, NBV selection is typically solved in a greedy manner.
The most commonly adopted approach is discrete view selection. Many algorithms have been developed for active scanning and/or recognition of single objects~\cite{Krainin2011,Wu2014,Xu2016} and scenes~\cite{Low2006,Fan2016}.
Recently, an object-centric approach has been proposed for view planning in active scene scanning~\cite{liu2018}.
However, the problem becomes significantly harder when the goal is continuous view planning or camera trajectory optimization.
Deep learning has shown great potential on NBV problem with the recent excellent works~\cite{Kollar2008,hepp2018learn}. Our work has a different problem setting in that we consider continuous \emph{path} optimization, far beyond discrete NBV prediction, for the task of dense scanning of large-scale outdoor environments.

\paragraph{Topological map for robot navigation}
Topological mapping has been a long-standing research topic in robotics. In the context of Simultaneous Localization and Mapping (SLAM), it is a common practice to partition an environment into a number of disjoint regions for the purpose of topological localization~\cite{badino2011visual}, hierarchical bundle adjustment~\cite{lim2012online}, or map reduction~\cite{dymczyk2015keep}. In~\cite{thrun1998learning}, the regions are delimited by narrow passages induced by the Voronoi decomposition of the environment. However, computing Voronoi diagrams on-the-fly for an unknown scene is difficult. Another type of approach to topological mapping is attaching local occupancy grids at different places along the metric SLAM map~\cite{konolige2011navigation}. Similarly, \cite{blochliger2018topomap} construct a topological map for robot navigation based on a convex decomposition of the occupancy grid. \cite{tang2019autonomous} employ a traditional frontier based method on the grid map of an outdoor scene, of which the construction relies on dense point clouds. Maintaining volumetric occupancy grids for outdoor scene scanning, however, is computationally intractable.

\begin{figure*}[t]
	\centering
	\begin{overpic}[width=\textwidth,tics=10]{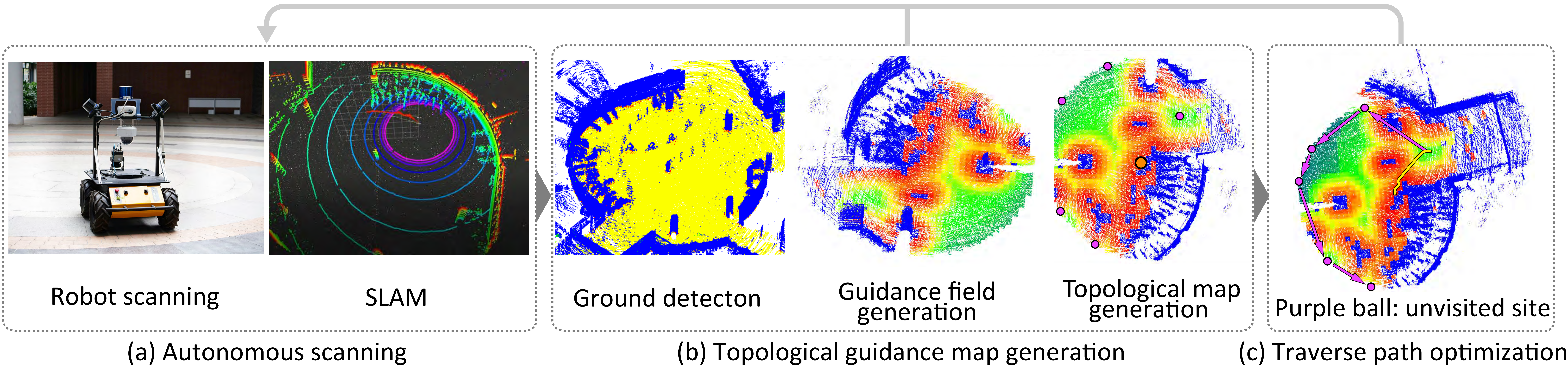}
	\end{overpic}
	\caption{An overview of proposed method.}
	\label{fig:overview}
\end{figure*}

\paragraph{TSP and Online TSP for robot navigation}
Robot exploration and navigation is inherently a TSP problem when the visiting goals are known to the robot.
There is a large body of prior works that formulate robot exploration as a TSP~\cite{sariel2005real,imeson2014language,kulich2014single,liu2019ground}, where the graph of visit sites is constructed before the robot moves. Multiple TSP (MTSP) has also been exploited to realize multi-robot collaborative mapping of scenes~\cite{Ortolf2012,sariel2005real}. When the topological map of an outdoor environment is unknown a priori, however, the problem becomes an Online TSP in which the robot needs to dynamically determine a set of sites to visit, and in which order, so that the total collected reward is maximized and a given time budget is not exceeded.
In the task of autonomous scanning, the sites are defined as goal positions to be visited to maximize the scan coverage as much as possible. Therefore, these sites are dynamically evolving as the scanning proceeds. We are not aware of any work that formulates robot-operated autonomous scanning with Online TSP, where one needs to handle on-the-fly insertion and deletion of scanning-driven sites like what we solve.

\section{Overview}
\label{sec:overview}

\paragraph{Problem, objective and formulation}
The goal of this work is to drive a mobile robot, mounted with a laser scanner,
to explore and scan an unknown environment with on-the-fly planned scanning paths,
to maximally cover the scene with the minimal amount of robot movement (energy consumption).

As shown in \Fig{overview}, we formulate the on-the-fly planning as an interleaving, discrete-continuous
optimization of robot scanning paths, within an online, progressively constructed exploration map.
The discrete optimization constructs a topological guidance map with dynamically
determined scanning goals, and solves an Online TSP over the topological map to obtain an optimal scanning path. The continuous part optimizes the traverse path between every two
sites visited consecutively along the TSP path, through maximizing
the information gain of scanning along it.

\paragraph{Topological guidance map}
To guide an efficient exploration and scanning,
we construct a topological map to direct the robot into those explorable regions, which are more visible to the unknown part of the environment.
To this end, we first compute a \emph{point-based guidance field} over the points scanned so far.
This point-based map encodes the \emph{explorability} and \emph{observability} of each point.
Given a point, its explorability is defined as the distance of the point to the robot,
as well as the distance to the closest point on the local medial axis defined by
the surrounding boundaries/objects around the point being considered.
The observability (against unseen regions) of a point is measured via estimating how much unseen regions could be revealed if the robot moved to that point.
Based on the point-based field, we can sample a set of local maximal points out of the current point cloud,
forming the visit sites of our topological guidance map over which the traverse path is computed.

\paragraph{Traverse path optimization}
After solving for the TSP path over the on-the-fly constructed topological map,
we refine the traverse paths through maximizing the information gain of scanning along the path.
Given a traverse path between two adjacent visit sites,
we define the information gain as the amount of scanning uncertainty visible to that path.
Given an acquired point, its local scanning quality and uncertainty are measured based on the vicinity of that point. The information gain can then be estimated at any given point.
Based on the estimated information gain, we optimize a B-spline curve around
the shortest traverse path between two sites to obtain the final scanning path.

\section{Discrete, Topological Path Optimization}
\label{sec:discrete}

The goal of our path optimization is to maximize the point cloud acquisition during the explorative scanning while minimizing the robot moving distance:
\begin{equation}\label{eq:op-energy}
    \min.\sum_{\{s_i,s_j\}\in \mathcal{T}} {\frac{d(s_i, s_j)}{\Omega(s_i, s_j)}},
\end{equation}
where $s_i$ is a visit site of the robot, $d$ measures the length of a path segment $(s_i, s_j)$, and $\Omega$ represents the area of surface that can be scanned along the path segment. We will elaborate this objective via formulating it as an online TSP.

\subsection{Topological guidance map}
\label{sec:topomap}
The general idea of online path planning is to guide the robot exploration
through analyzing the online acquired data.
An important feature of LiDAR acquisition, i.e., \emph{$360^\circ$ round scanning} with the elevation angle ranging in $[-15^\circ,15^\circ]$,
enables us to construct a \emph{local} guidance map that evolves dynamically as the robot moves,
to emulate a \emph{global} guidance map directing the robot exploration.

\begin{figure}[t]
\centering
\begin{overpic}
[width=0.88\linewidth]
{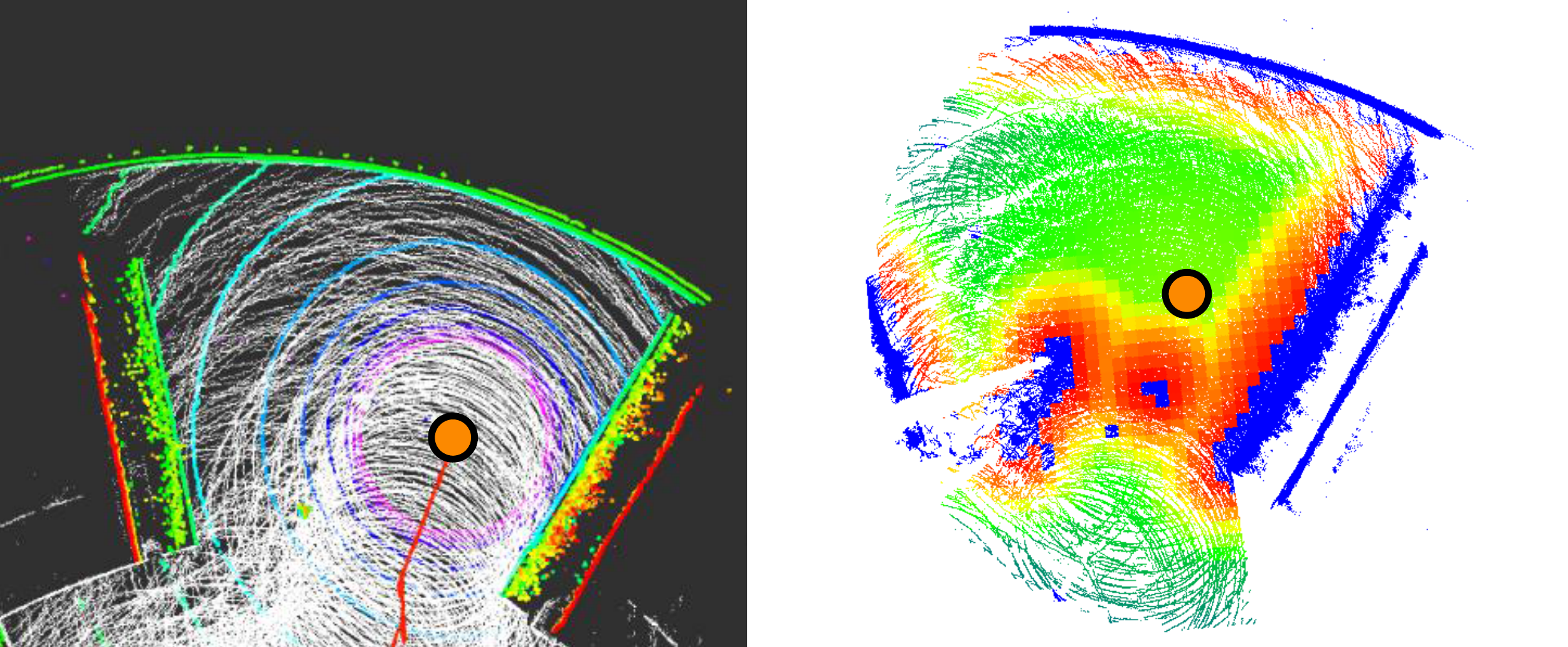}
\myfigurename{}
\end{overpic}
\caption{
Left: 360$^\circ$ round scans made by the LiDAR scanner, centered at the robot location (orange point).
Right: the guidance field and topological map are both estimated over the point cloud within a round region of a specific radius around the scanning robot.
}
\label{fig:roundscan}
\end{figure}

At each scanning position, a LiDAR scanner performs a $360^\circ$ scanning within a valid range,
forming a round-ranged point cloud centered at the robot location. Based on the point cloud acquired so far, we construct a \emph{topological} guidance map within a circular area
around the robot with a specific radius (see \Fig{roundscan}). We refer to this circular area as
the \emph{planning area} of a given robot location.
As the robot moves forward, the topological map is updated based on the point cloud around the new robot position,
after integrating the newly acquired points. This ensures that there is always a dynamically updated
guidance map around the robot, as if there was a global, underlying guidance map.

To construct the topological map, we first compute a point-based guidance field over the point
cloud within the current planning area. The point-based field is estimated over only the points representing
the ground. For each ground point, we estimate its explorability for the robot as well as
its observability against unseen regions if the robot was standing at the point.
Next, we elaborate the computation of point-wise explorability and visibility.

\paragraph{Point-wise explorability}
Given a point, its explorability encompasses two aspects.
First, to encourage robot exploration, we assign high explorability for the points distant from
the robot location. Specifically, the \emph{distance-based explorability} of a 3D point $p$ is:
\begin{equation}\label{eq:traverse-d}
  \phi_\text{d}(p,p_\text{r}) = 1 - e^{-\frac{\|p-p_\text{r}\|^2}{\sigma^2}},
\end{equation}
where $p_\text{r}$ is the location of the robot.
In general, $\sigma$ takes the maximum valid range of LiDAR scanning. We will study the effect of this parameter in \Sec{results}.

Second, the acquisition of $360^\circ$ scanning is most efficient when the scanner is placed in
a spacious region, thus avoiding occlusions as much as possible.
To direct the robot into relatively spacious regions while avoiding obstacles,
we consider a point to be more explorable if it is far away from obstacles (e.g., walls and objects).
Inspired by the definition of L1 medial axis for point clouds~\cite{huang2013l1},
we define the \emph{medial-based explorability} of a point as:
\begin{equation}\label{eq:traverse-m}
  \phi_\text{m}(p) = \min\{1, \frac{1}{\sigma}\min_{o\in\mathcal{O}}\|p-o\|\},
\end{equation}
where $\mathcal{O}$ is the point set representing the obstacles surrounding point $p$.
It measures how close the point is to the local medial axis defined by its surroundings.
The larger $\phi_\text{m}$ is, the emptier the space around the point is,
and consequently the more explorable to the robot the point is.

Combining these two measures, we obtain the integrated explorability:
\begin{equation}\label{eq:traverse}
  \phi(p,p_\text{r}) = \omega\phi_\text{d}(p,p_\text{r}) + (1-\omega)\phi_\text{m}(p),
\end{equation}
with $0<\omega<1$ being the weight tuning the importance of the two measures.
We set $\omega=0.5$ throughout our experiments.
As the robot moves, some points will be observed multiple times.
Therefore, for the points whose explorability is measured multiple times,
only the minimum value is kept.
This way, the explorability of any point decreases monotonically as the scanning proceeds.
This feature facilitates an efficient scene exploration through minimizing tedious back-and-forth visits.

\paragraph{Point-wise observability}
The key to guide robot exploration in an unknown environment is to drive it to quickly discover the unknown regions. A classical approach to this is frontier-driven~\cite{Thrun2005},
where the environment is represented with a 2D grid or 3D volume.
A grid cell (or voxel) can be either empty, occupied, or unknown, depending on whether
it is known to be occupied by an obstacle based on robot observation.
Frontiers are the cells lying in the interface between empty and unknown cells.
Driving the robot to explore frontiers would encourage it to discover more unknown regions.
We find it difficult, however, to adopt the frontier-based exploration in our problem setting.
First, maintaining a volumetric representation for an outdoor open area is neither memory efficient
nor computationally tractable. Second, the sparse nature of laser-based range finder, especially
for far ranges, makes it hard to correctly compute the occupancy status of a grid cell.

\begin{figure}[t]
\centering
\begin{overpic}
[width=0.88\linewidth]
{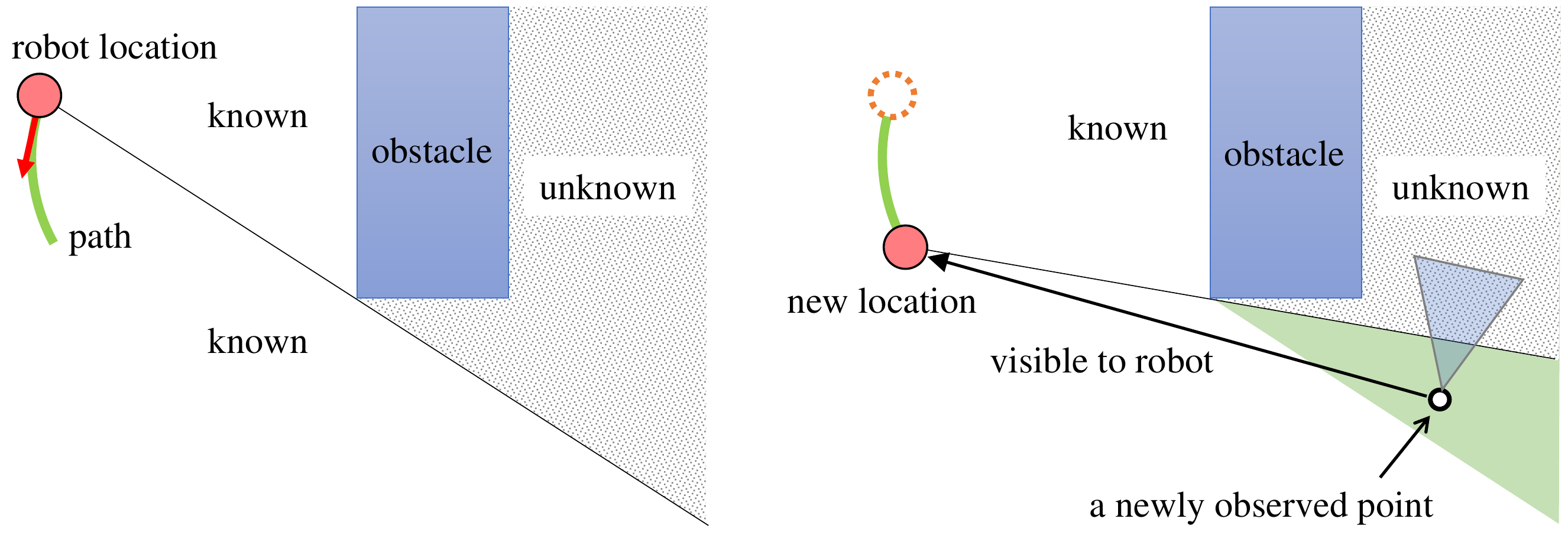}
\myfigurename{}
\end{overpic}
\caption{
Illustration of observability estimation based on newly observed points.
Left: the original robot location and the corresponding seen and unseen regions due to the occlusion by the obstacle.
Right: after the robot moves to a new location, the green region becomes visible to the robot. Any point in that region is \emph{newly observable} to the robot since the last time step. Such points have high observability to the unknown regions caused by the same obstacle, due to visibility continuity.
}
\label{fig:observability}
\end{figure}

To avoid representing the whole environment, we opt to estimate unknown regions
based on the already scanned point cloud. This option is seemingly unrealistic at first sight:
The already scanned points have to be known, how can one estimate unknowns over these known points?
Our key observation supporting this idea is that \emph{the exposure of unknown regions due to occlusion
is typically continuous, as the observer moves around the obstacles}; see \Fig{observability}.
This makes it possible to \emph{hallucinate} unknown regions, via estimating
the \emph{observability-to-unknown} of an ever-observed point. The latter is achieved by judging
whether the point is newly observed by the robot w.r.t. the historical locations of the robot within a past time interval.
If it is newly observed, the point would be of high interest to the robot since moving towards
it could potentially expose more unknown, based on the assumption of continuous occlusion.

Consequently, a point $p \in \mathcal{P}$ has an observability-to-unknown if it is visible to
any of the historical robot locations (viewpoints), $p^t_\text{r}$ ($t=t_c-k,\ldots,t_c-1$),
in the past $k$ time steps back from the current time $t_c$ (we set $k=2$ by default).
Formally, the observability of point $p$ is defined as:
$\rho(p, p^{t_c}_\text{r}) = 1-\prod_{t=t_c-k}^{t_c-1}{(1-\theta(p, p_\text{r}^{t}))}$,
where $\theta(p,p_\text{r})$ is an indicator function indicating whether a point $p$ is visible
w.r.t. the robot location $p_\text{r}$.
Our next task is then to determine the visibility of a 3D point against a given viewpoint.

In determining point cloud visibility,
we hope to avoid both surface reconstruction which is too costly for online planning,
and normal estimation which is intractable for the sparse and noisy LiDAR point cloud.
We therefore adopt the Hidden Point Removal (HPR) operator~\cite{katz2007direct}, which is an elegant method
of direct visibility determination for 3D point clouds with theoretical guarantees.
The operator is very simple: It first transforms the 3D points to a new domain and then computes the convex
hull of the transformed points in that domain. Points that lie on the convex hull then correspond to the visible
points in the original point cloud. In this work, we utilize its enhanced variant, i.e., the generalized HPR operator (GHPR)~\cite{katz2015visibility}.

Given the point set $\mathcal{P}$ for which we would like to determine the visibility against a viewpoint at $p_\text{r}$,
we transform each point $p \in \mathcal{P}$ and $p_\text{r}$ using the following radial transformation:
\begin{equation}\label{eq:visibility-transform}
  f(p,p_\text{r}) = \left\{
\begin{array}{lcl}
p_\text{r} + \frac{p-p_\text{r}}{\|p-p_\text{r}\|}h(\|p-p_\text{r}\|),       &      & p \neq p_\text{r}\\
p,     &      & p=p_\text{r}
\end{array} \right.
\end{equation}
where $h$ is a 1-dimensional continuous kernel function:
$$
h(\|p-p_\text{r}\|) = \alpha \max_{q\in \mathcal{P}}(\|q-p_\text{r}\|)-\|p-p_\text{r}\|,
$$
where $\alpha=10^4$ is a scaling factor.
Based on the convex hull of the transformed point set, we can find the visible set w.r.t. $p_\text{r}$.

\begin{figure}[t]
\centering
\begin{overpic}
[width=0.88\linewidth]
{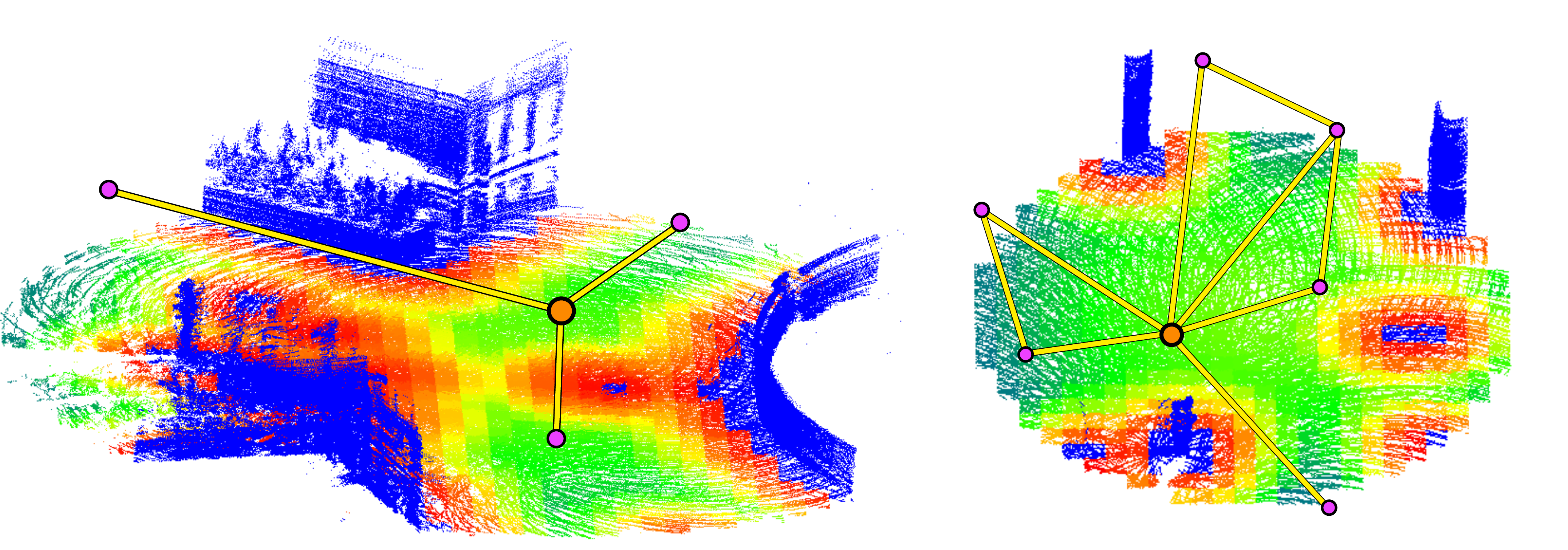}
\myfigurename{}
\end{overpic}
\caption{
Two examples of topological guidance map. The orange dots represent the current robot location, while the purple
dots denote the visit sites. The topological map is formed by a $k$NN graph of all sites.
}
\label{fig:topomap}
\end{figure}

\paragraph{Topological map construction}
Having obtained the point-based explorability and \obf, our next step is to combine them to form
a point-wise guidance field, from which a topological map can be extracted.
Since the \obf is binary, we smooth it through convolving it with a Gaussian kernel $g(\cdot)$ at each point, i.e.,
$\tilde{\rho}(p, p_\text{r}) = \rho(p, p_\text{r}) \circledast g(p)$.
The combined guidance field is then defined by:
\begin{equation}\label{eq:guidemap}
  \tau(p, p_\text{r}) = \lambda \phi(p,p_\text{r}) + (1-\lambda) \tilde{\rho}(p, p_\text{r}),
\end{equation}
where we set $\lambda=0.7$ by default. The high value regions in the point-based guidance field
correspond to those traversable and worth-to-explore.
We extract the local maximal points of the field with non-maximum suppression (NMS),
leading to a collection of visit sites for the current robot location:
\begin{equation}\label{eq:visitsite}
\mathcal{S}(p_\text{r}) = \{s_i | s_i=\argmax_{p\in N(s_i, r_\text{n})}(\tau(p, p_\text{r}))\},
\end{equation}
where $N(s_i, r_\text{n})$ is the neighborhood of site $s_i$ within a radius of $r_\text{n}$.
The selection of the radius $r_\text{n}$ depends on the level of details of scanning.
A fine-grained scanning requires a small $r_\text{n}$; we use $r_\text{n}=5$m in all our experiments (see \Sec{results} for a study on its selection).
The topological guidance map is then constructed as the $k$NN graph ($k=8$) of the visit sites.
\Fig{topomap} shows two examples of topological guidance map built upon the scanned point cloud for
the current robot location.
It can be observed that our method tends to place visit sites at the portals between
different connected passages, which is natural for navigating the robot's exploration.

\subsection{Path optimization as an Online TSP}
\label{sec:otsp}
Based on the topological guidance map, our next step is to compute a traverse path
for the robot to visit the sites in the map while minimizing the energy consumption.
A straightforward solution seems to formulate the problem as a Traversing Salesman Problem (TSP).
However, the underlying point-based guidance field changes on-the-fly during the traversing and scanning. Accordingly, the topological map needs to be updated through
inserting newly discovered sites and deleting visited or well-observed sites.
Planning robot paths over such dynamically changing map calls for a dynamic variant of TSP,
for which we employ the formulation of Online Traveling Salesman Problem (Online TSP). In Online TSP, both the sites to visit and the order of visiting are determined online to minimize
the total traverse cost.
To our knowledge, our work is the first that applies Online TSP in robot-operated autonomous scanning.

\paragraph{Objective formulation}
Given the current set of visit sites $\mathcal{S}=\{s_i\}_{i=0}^{S}$,
we define the reward of visiting a site $s_i$ from site $s_j$ as the information gain of scanning along the traverse path.
The traverse cost between two sites $s_i$ and $s_j$ simply takes the shortest distance between them: $d(s_i, s_j)$.
The optimal traverse path $\mathcal{T}$ can be found through minimizing the following objective:
\begin{equation}\label{eq:op-energy}
  \min. \sum_{\{s_i,s_j\}\in \mathcal{T}} {\frac{d(s_i, s_j)}{1-\tau(s_i, p_r)}},
\end{equation}
where $\{s_i,s_j\}$ is a segment in path $\mathcal{T}$ and $\tau(s_i, p_r)$ is the
guidance field value (\Eq{guidemap}) at site $s_i$ given the robot location $p_r$.

\paragraph{Optimization}
The optimization of Online TSP involves on-the-fly determination of visit sites and visiting order.
In determining the visit sites, our policy is to first quickly visit the sites in
open area and then traverse those which are located near a branch or a cross road.
The former simply takes a greedy scheme and the latter involves solving an TSP.
Therefore, we classify the sites into two types. A site is an Open Area (OA) site if its
medial-based explorability $\phi_\text{m}(p)>0.8$ or observability $\tilde{\rho}(p, p_\text{r}) = 1$,
and a Branch Entry (BE) site otherwise.
The path optimization process is as described in \Algo{gtr}.
For each newly coming scan, site update happens only within its scan range. In particular, the guidance field at each site is reevaluated (Line 6), and those whose value is lower than a threshold (0.3 by default) are removed.
If there are OA sites, the robot simply visits them greedily (Line 11).
Then for the remaining BE sites, an optimal path is computed by solving Online TSP based on Mixed Integer Programming (Line 13), optimizing the objective in \Eq{op-energy}.
After the robot moves, new sites are generated and added into the active set, based on the updated guidance field (Line 3-4).

\begin{algorithm}[t]
	\SetCommentSty{textsf}
	\SetKwInOut{AlgoInput}{Input}
	\SetKwInOut{AlgoOutput}{Output}
	\SetKwFunction{Scanning}{Scanning}
	\SetKwFunction{Guidance}{CompGuidanceField}
	\SetKwFunction{GenNodes}{GenerateSites}
     \SetKwFunction{Classify}{ClassifySites}
     \SetKwFunction{Greedy}{GreedyVisit}
     \SetKwFunction{Maintain}{MaintainActiveSet}
     \SetKwFunction{MIP}{TSPVisit}
	\SetKwFunction{RewardCost}{CompReward\&Cost}
	\SetKwFunction{Objective}{Objective}
	\AlgoInput{ Initial position of robot: $p_r$ (init.: $p_\text{r} \leftarrow p_0$)}
	\AlgoOutput{ Scanned point cloud: $\mathcal{P}$ (init.:  $\mathcal{P} \leftarrow \varnothing$)}


     \Repeat {$\mathcal{S}_\text{active}$  $\not=  \varnothing$} {

		$\mathcal{P}$ $\leftarrow$ \Scanning{$p_\text{r}$}\;
	
		$\tau(p_\text{r})$ $\leftarrow$ \Guidance{$\mathcal{P}$, $p_\text{r}$}\;
		
		$\mathcal {S}(p_\text{r})_=\{s_i\}_{i=1}^{S}$ $\leftarrow$ \GenNodes{$\tau(p_\text{r})$}\;

           $\mathcal \{{S}_\text{OA}(p_\text{r}),{S}_\text{BE}(p_\text{r})\}$ $\leftarrow$ \Classify{${S}(p_\text{r})$}\;

           $\mathcal{S}_\text{active}$ $\leftarrow$ \Maintain{$\mathcal{S}_\text{active} = \{{S}_\text{OA} \cup {S}_\text{BE}\}$}\;
		

          $\mathcal{S}_\text{OA} \leftarrow {S}_\text{OA} \cup {S}_\text{OA}(p_\text{r})$ and $\mathcal{S}_\text{BE} \leftarrow {S}_\text{BE} \cup {S}_\text{BE}(p_\text{r})$\;
					
		$\{r(s_\text{i},s_\text{j}),d(s_\text{i},s_\text{j})\}_{i=1,j=1}^{S}$ $\leftarrow$ \RewardCost{$\mathcal{S}_\text{active}$}\;
			
		$\{o(s_\text{i},s_\text{j})\}_{i,j=1}^{S}$ $\leftarrow$ \Objective{$\{r(s_\text{i},s_\text{j}),d(s_\text{i},s_\text{j})\}_{i,j=1}^{S}$}\;
			
	     \eIf{$\mathcal{S}_\text{OA}$ $\not= \varnothing$ }{

               $s_\text{best}(p_\text{r})$ $\leftarrow$ \Greedy{$p_\text{r}$,$\{o(s_\text{i},s_\text{j})\}_{i,j=1}^{S}$}\;

          }{
               $s_\text{best}(p_\text{r})$ $\leftarrow$ \MIP{$p_\text{r}$,$\{o(s_\text{i},s_\text{j})\}_{i,j=1}^{S}$}\;
          }
	
		$p_r \leftarrow s_\text{best}(p_\text{r})$ \;
     }
\caption{Path Optimization}
\label{algo:gtr}
\end{algorithm}

\section{Continuous, Geometric Path Optimization}
\label{sec:continuous}
After obtaining a TSP path over the topological map, the final traverse path is computed
by estimating a geometrically continuous path between every two consecutively visited sites.
A natural option for this is to compute the shortest path between the two sites over the $k$NN graph of acquired points connecting the two sides.
To make the traverse paths more aware of scan quality, we perform geometric refinement to make them pass
through the regions needing more scans as much as possible, resulting in \emph{quality-aware scanning paths}.
In achieving that, we first need a method to assess the scan quality of the point cloud.
The path refinement is then conducted based on the quality assessment.

\subsection{Shape-aware point cloud quality assessment}
The quality evaluation of 3D point clouds is still an open problem
despite the long-standing development of laser scanning and point-based representation~\cite{torlig2018novel}.
Most existing approaches require the availability of a reference point cloud and perform quality evaluation
based on point cloud comparison~\cite{alexiou2017performance}.
For the task of autonomous scanning of objects by a robot, \cite{Wu2014} propose to
measure the quality of a point cloud based on the gradient of the Poisson field fitting the point cloud.
A probabilistic metric based on volumetric representation of point clouds is proposed in~\cite{Krainin2011} for the same task.
These metrics are unsuitable for our task since the computation of volumetric reconstruction for
the point cloud of outdoor scenes is computationally intractable.
A purely density-based measure is unsuited either for two reasons.
First, the density of LiDAR data is anisotropic so the measuring is difficult.
Second, we hope our scanning to be aware of the target object.
For example, dense scanning should be devoted to walls or building facade but not to vegetation.
Therefore, a \emph{shape-aware quality measure} is demanded.

\paragraph{Fractal dimension}
In fractal geometry, a fractal dimension is an index for characterizing fractal patterns or sets by quantifying their complexity as a ratio of the change in detail to the change in scale.
It has been extensively used in medical image analysis as a morphological characteristics of tissue.
We are not aware of a work on using fractal dimension to assess the quality of 3D point clouds.
In fact, fractal dimension is especially suitable for the task due to its rotation- and scaling-invariance,
as well as noise- and occlusion-insensitivity.

\begin{figure}
\centering
\begin{overpic}
[width=\linewidth]
{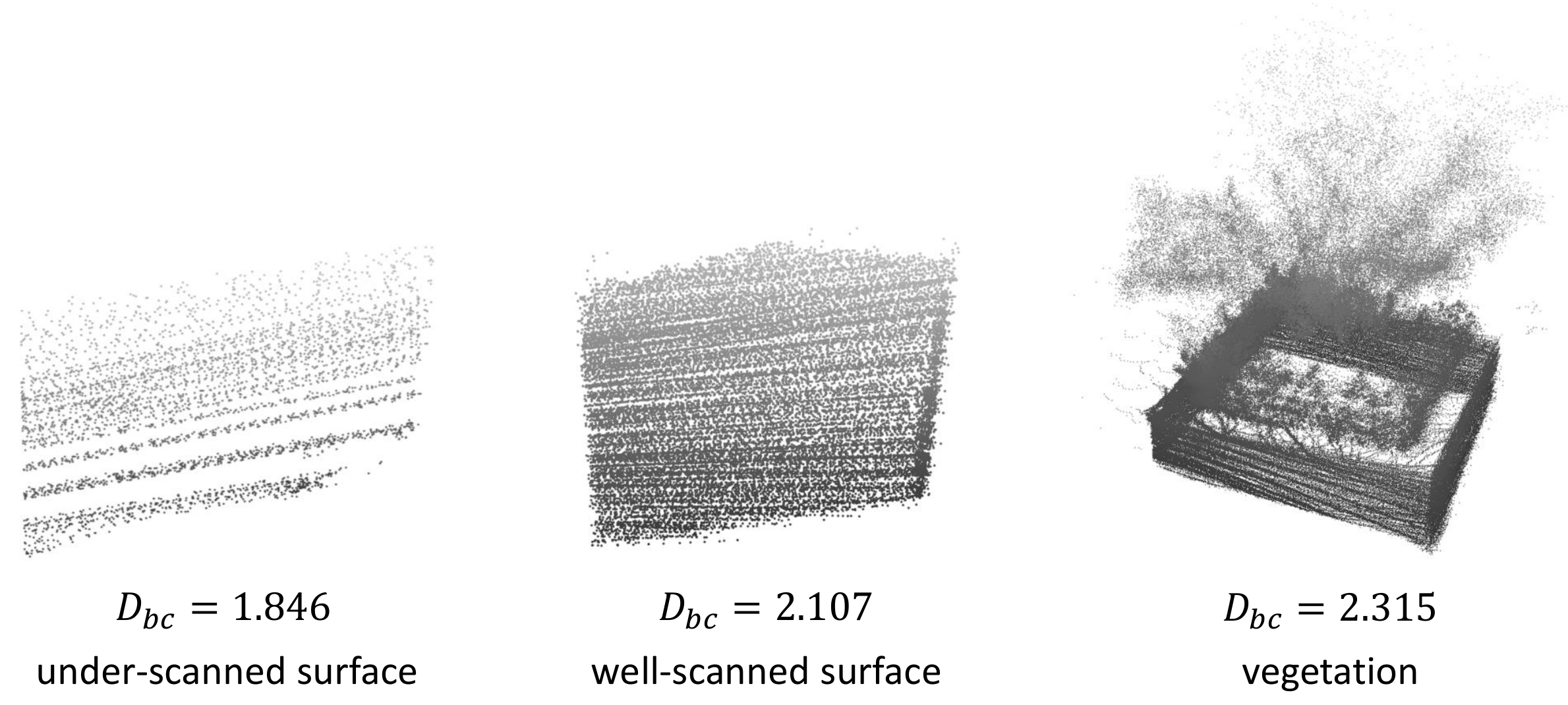}
\end{overpic}
\caption{Fractal dimensions of different point sets: an under-scanned surface (left),
a well-scanned surface (middle), and a vegetation area (right).}
\label{fig:fractal}
\end{figure}

For a point set describing a geometric shape, the fractal dimension represents the Euclidean or topological dimension of the shape as introduced in~\cite{traina2010fast}. For example, sets of dimension $0$, $1$, $2$ and $3$ describe points, lines, surfaces and volumes, respectively.
\cite{yang2015extraction} utilize fractal dimension analysis to extract vegetation from 3D point clouds.
Inspired by this, we assess the local quality of a point cloud using fractal dimension analysis. Given a point in a scanned point cloud, we estimate the fractal dimension for the point set around it, to determine the characteristics of that point.
As shown in \Fig{fractal}, with sufficient scanning, the scanned point clouds, representing the surface of the scene (ground and obstacles), should have a fractal dimension of $2$.
Consequently, point sets with a fractal dimension smaller than $2$ can be regarded either as outliers or under-scanned parts. On the contrary, point sets whose fractal dimension is higher than $2$ may describe structurally complex objects, such as vegetation.

\begin{figure}
\centering
\begin{overpic}[width=0.45\linewidth]
{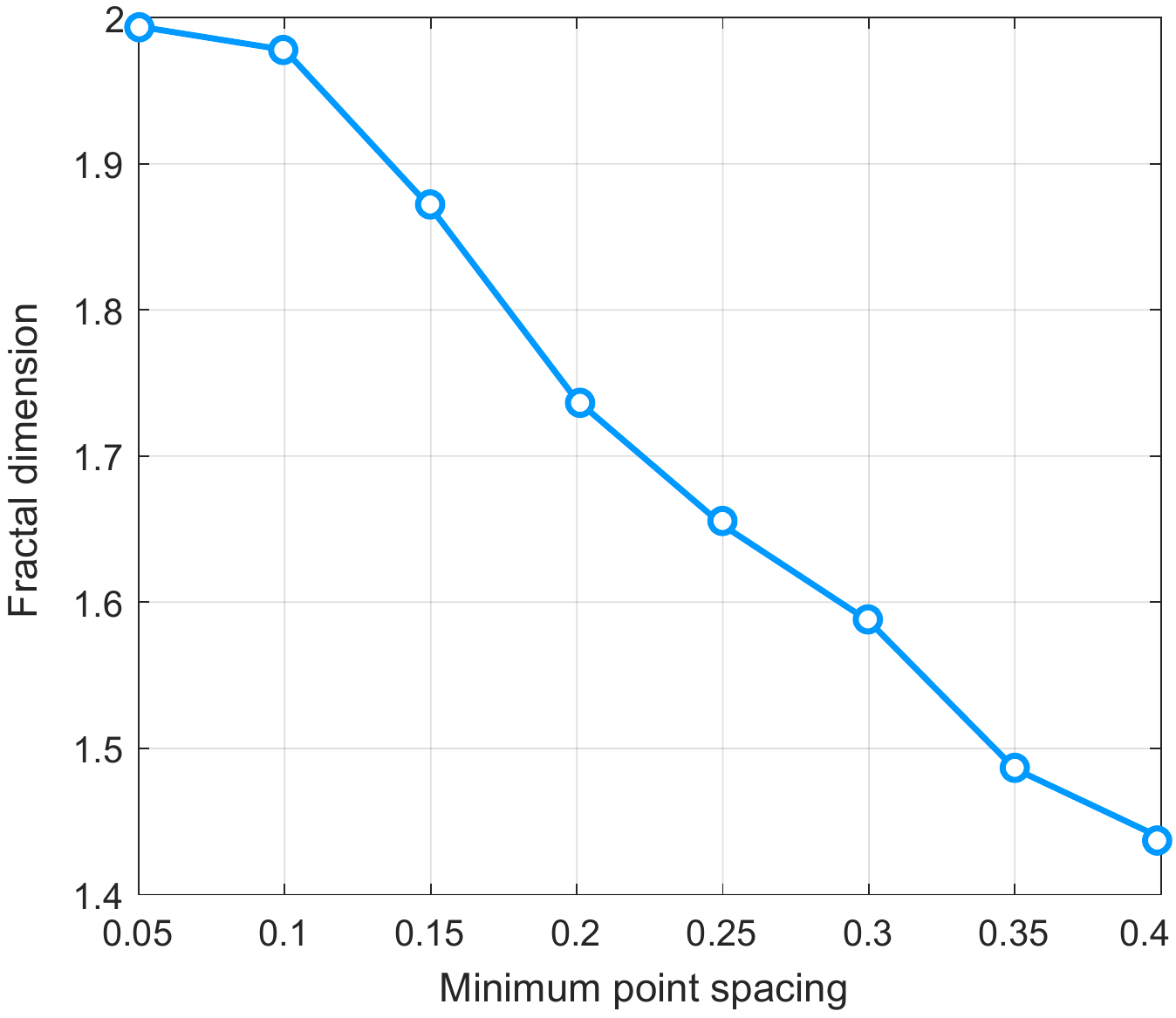}
\end{overpic}
\begin{overpic}[width=0.45\linewidth]
{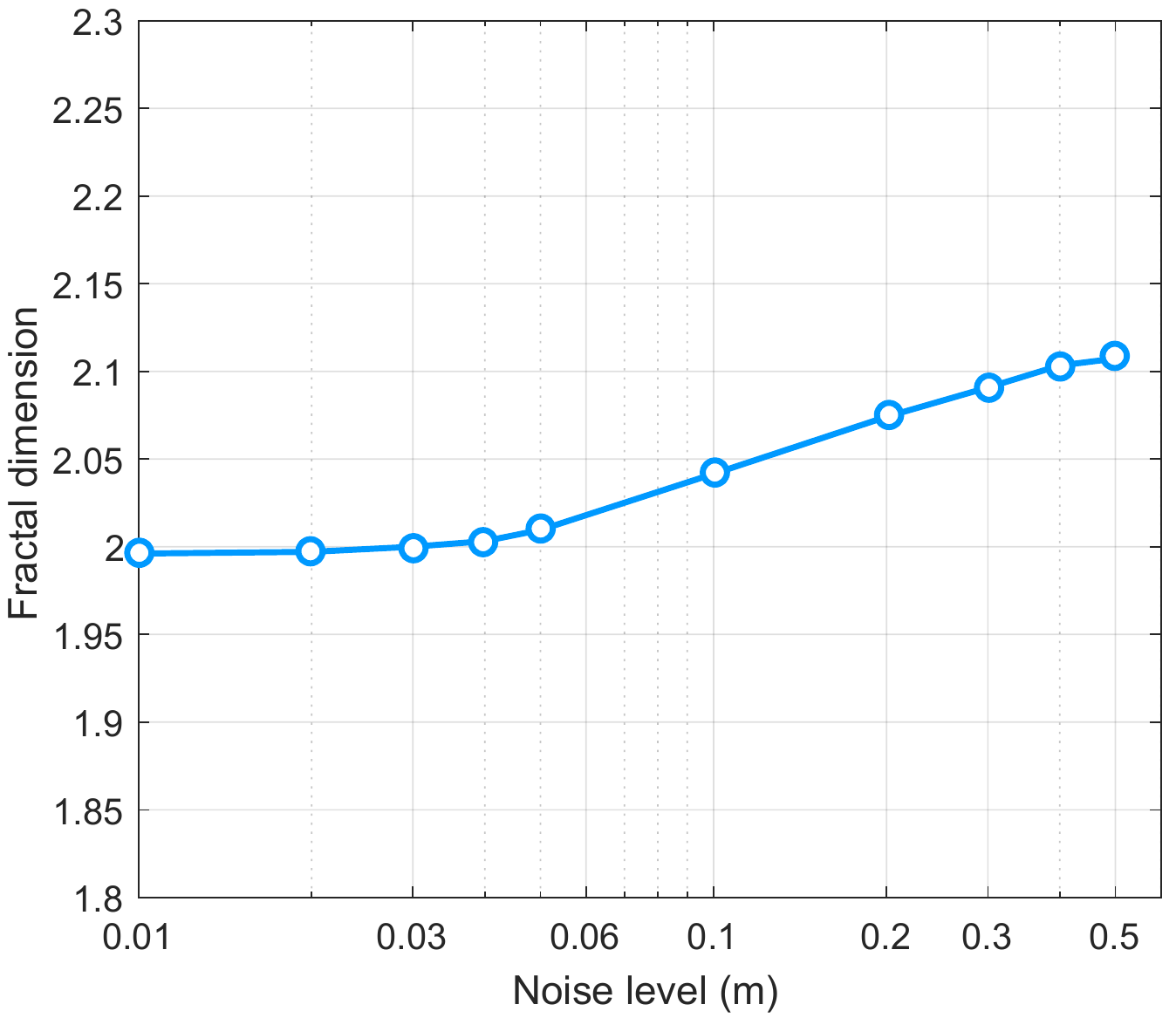}
\end{overpic}
\caption{
Fractal (box-counting) dimensions of point clouds with decreasing sampling rate (left)
and with increasing noise level (right).
}
\label{fig:fractal_plot}
\end{figure}

\paragraph{Box-counting dimension}
To computationally realize the fractal dimension analysis, we utilize the box-counting dimension (a.k.a., Minkowski-Bouligand dimension)~\cite{falconer2004fractal}, which is a grid-based discretization of fractal dimension computation.
Given a 3D point set $\mathcal{P}$, we embed it into a volumetric grid and count how many grid cells (boxes) are required to cover the set. The box-counting dimension is then calculated by measuring how this number changes as we increase the resolution of the grid. Suppose that $C(\mathcal{P},\epsilon)$ is the number of grid cells of size $\epsilon$ required to cover the set $\mathcal{P}$. The box-counting dimension is defined as:
\begin{equation}\label{eq:bc-dim}
D_\text{bc}(\mathcal{P}) = \lim_{\epsilon\rightarrow 0}{\frac{\log C(\mathcal{P},\epsilon)}{\log(1/\epsilon)}}.
\end{equation}
In practice, we collect a series of pairs $\{\log C(\mathcal{P},\epsilon), \log(1/\epsilon)\}$ for a list of increasing $\epsilon$'s, perform a least-square fitting to these data, and compute the gradient of the fitted line as an approximation of the box-counting dimension. The increasing resolutions of volumetric grid can be efficiently implemented with the help of octree-based representation. \Fig{fractal_plot} shows how fractal dimension can be used to measure the sparseness of the scanning surface. Moreover, the right subfigure shows that fractal dimension is robust to noise, making it appropriate to handle raw LiDAR data.

\subsection{Quality-driven path refinement}

To direct the robot into the regions that need more scans, we reform the shortest path between the current
and the next site into a B-spline curve passing through more under-scanned points.
This is achieved by selecting the control points of the B-spline curve as those points with
low fractal dimension (i.e., under-scanned).
Specifically, from the point cloud scanned at the current robot location, we first remove
those whose fractal dimension is equal to or larger than $2$. We then sort the remaining points
into a priority queue, with the ascendant order of fractal dimension.

\begin{figure}
\centering
\begin{overpic}
[width=\linewidth]
{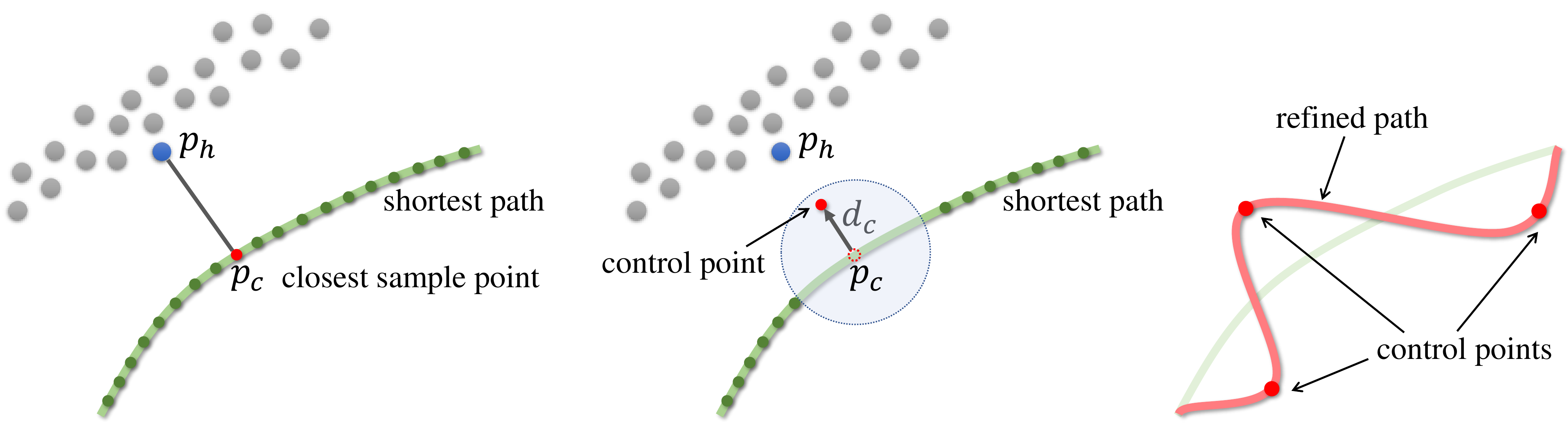}
\end{overpic}
\caption{
Illustration of quality-driven path refinement. Given the point $p_\text{h}$ from the head of the quality queue,
the closest sample point $p_\text{c}$ on the shortest path is found (left). Then $p_\text{c}$ is moved towards
$p_\text{h}$ by a distance $d_\text{c}$, forming a control point; all sample points within a circle (with a radius of $r_\text{c}$) centered at $p_\text{c}$ are removed (middle). The process repeats until there is no more sample point left on the shortest path. Finally, a B-spline curve is computed based on the selected control points (right).
}
\label{fig:pathrefine}
\end{figure}

During control point selection, we choose a point $p_\text{h}$ from the head of the queue and search
for the closest point $p_\text{c}$ on the point sampled shortest path. We then move $p_\text{c}$ towards $p_\text{h}$
for a prescribed distance $d_\text{c}=1.5$m. This moved point is then selected as a control point. After that, $p_\text{h}$ is removed from the priority queue and any points on the shortest path, whose distance
to $p_\text{c}$ is smaller than a given radius $r_\text{c}$, are removed. The process is repeated until there is no more sample
point left on the shortest path. \Fig{pathrefine} shows an illustration of the process.
Finally, we compute a third-order B-spline based on the selected control points, leading to the refined
traverse path in replace of the shortest path.
This smooth path can significantly reduce the sudden deflection of the robot motion, and thus alleviates SLAM deviation.

\section{Implementation}
\label{sec:impl}

\begin{figure}[t]
\centering
\begin{overpic}
[width=\linewidth]
{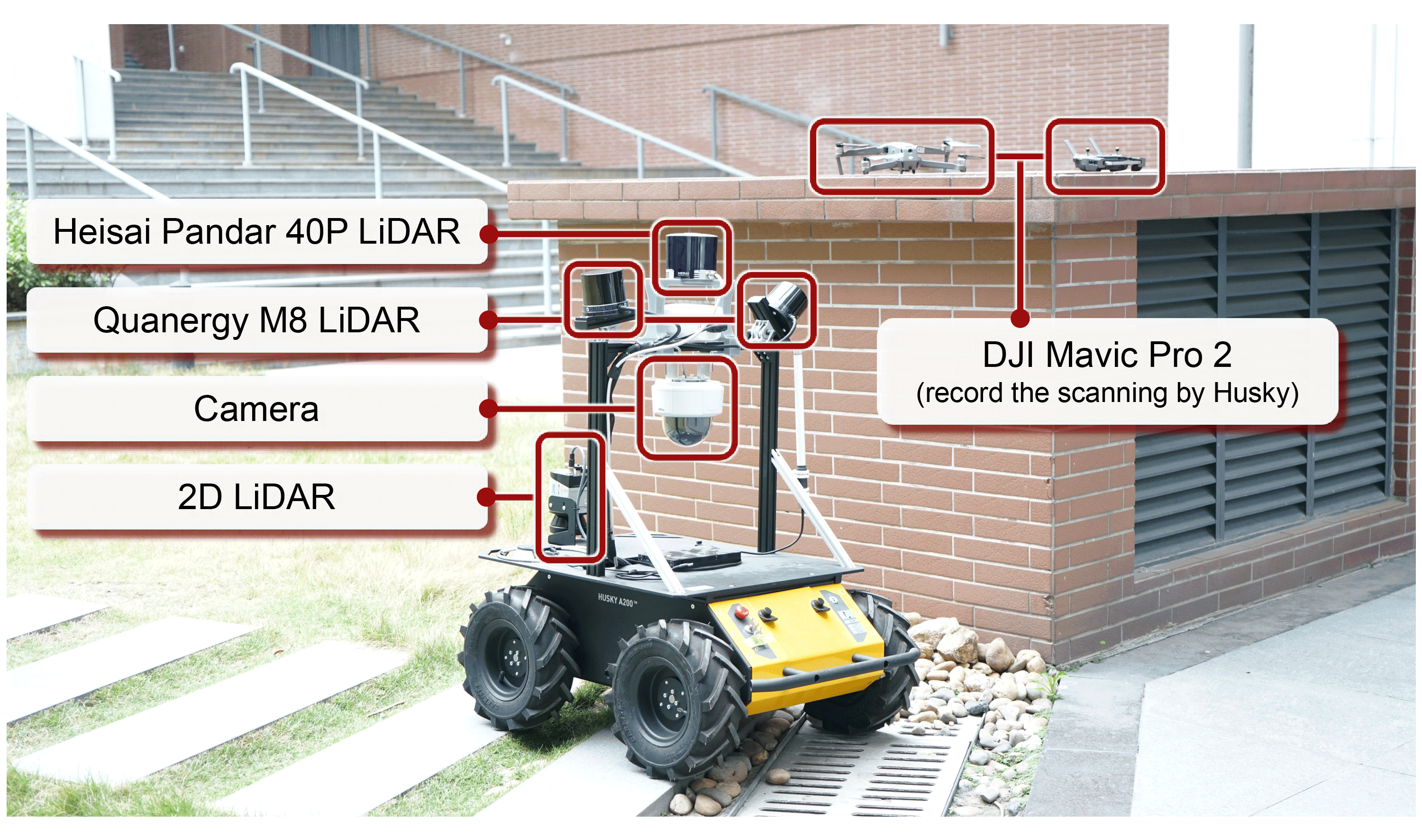}
\myfigurename{}
\end{overpic}
\caption{
Our auto-scanning system, a Husky robot equipped with a variety of sensors, among which only the Hesai Pandar 40P LiDAR sensor is used. We also utilized a UAV for video recording of the scanning process.
}
\label{fig:system}
\end{figure}

\paragraph{System setup}
To realize our algorithm, we utilize a Husky robot equipped with one Hesai Pandar 40P LiDAR scanner,
two Quanergy M8 LiDAR scanners, and an RGB camera (see \Fig{system}).
We only use the Hesai Pandar 40P LiDAR for both SLAM, online path planning, and scene acquisition.
For robot localization, we utilize the LiDAR-based SLAM method LOAM~\cite{zhang2014loam}; no GPS or IMU based localization is employed.
Both SLAM and our online planning run on the carry-on computer (Intel(R) I5-4570TE CPU (2.7GHz$\times$4), 8GB RAM).
The battery life of the Husky robot is about $2$ hours. The LiDAR scanner is powered by an independent battery.

\paragraph{Ground extraction and collision avoidance}
In our implementation, the scanning area of an outdoor scene is defined as drivable area within $50$m distance from the start point of the robot. We utilize here the
Gaussian Process Incremental Sample Consensus (GP-INSAC) algorithm~\cite{douillard2011segmentation}
to detect the ground and obstacles. This method adopts an online learning process, which
performs segmentation for nearby point clouds based on a Gaussian regression model,
and then transfer the learned segmentation to the point clouds far away.
Based on detected ground and obstacles, our system drives the robot to avoid the obstacles
by at least $1$m when planning the traverse paths. This way, we do not need to maintain a
computationally costly occupancy map for collision avoidance.

\paragraph{Parameter setting}
The parameters in the Gaussian model of GP-INSAC are $\sigma_l=28$ and $\sigma_F=1.76$.
The thresholds involved in INSAC are $t_\text{model}=0.2$ and $t_\text{data}=0.8$.
Please refer to the original paper~\cite{douillard2011segmentation} for the meaning of these parameters.
The time step involved in computing observability is $8$ seconds.
The grid resolution for estimating box-counting dimension is set to $0.1$m.
The settings of all other parameters have been given in the technical sections.
\Sec{results} provides a study on the selection of the radius of circular planning area and
that of the neighborhood used for NMS-based site selection.

\paragraph{Termination criteria}
In an unknown outdoor area, the completion of scanning is hard to define.
In our implementation, the termination of scanning is determined by estimating
whether the remaining battery life could afford the next planned traverse path.

\begin{table}[bp]\centering
	\caption{
		Statistics of the six real-world outdoor scenes. For each scene, we report the total area our robot scanned and the total traverse distance and time for finishing the scanning.
	}
	\scalebox{1.18}{
		\setlength{\tabcolsep}{3mm}{
			\begin{tabular}{l||c|c|c}
				\hline
				Scene            & Total area & Travel dist. & Total time \\ \hline\hline
				R-scene 1        & $2384$ m$^2$   & $125$ m  & $2.68$ min \\ \hline
				R-scene 2        & $1592$ m$^2$   & $113$ m  & $2.71$ min \\ \hline
				R-scene 3        & $1215$ m$^2$   & $153$ m  & $3.88$ min \\ \hline
				R-scene 4      & $1708$ m$^2$   & $92$ m   & $1.98$ min \\ \hline
				R-scene 5          & $1005$ m$^2$   & $71$ m   & $2.53$ min \\ \hline
				R-scene 6         & $1297$ m$^2$   & $97$ m   & $1.56$ min \\ \hline
			\end{tabular}
	}}
	\label{tab:stats}
\end{table}

\paragraph{Complexity}
The major computational cost for topological map generation is on the estimation
of medial-based explorability in \Eq{traverse-m}, which adopts KD-tree based spatially nearest point query.
The complexity of the KD-tree search is $O(m\log(n))$ with $m$ being the number of ground points and $n$ that of obstacle points.
The complexity of the observability estimation based on the GHPR algorithm is $O(n\log(n))$ with $n$ being the number of points in a planning area.
The complexity of TSP optimization is $O(2^n)$, where $n$ is the number of sites in the topological map.
Since the number of sites being considered by TSP is usually quite small ($<10$), the time for each TSP solving is typically $0.1$ second.
See \Sec{results} for more details.
The computational cost for fractal dimension estimation is $O(n\log(n))$ with $n$ being the size of the point set.

\section{Results and evaluations}
\label{sec:results}

\begin{figure*}[bht]
\centering
\begin{overpic}
[width=\textwidth]
{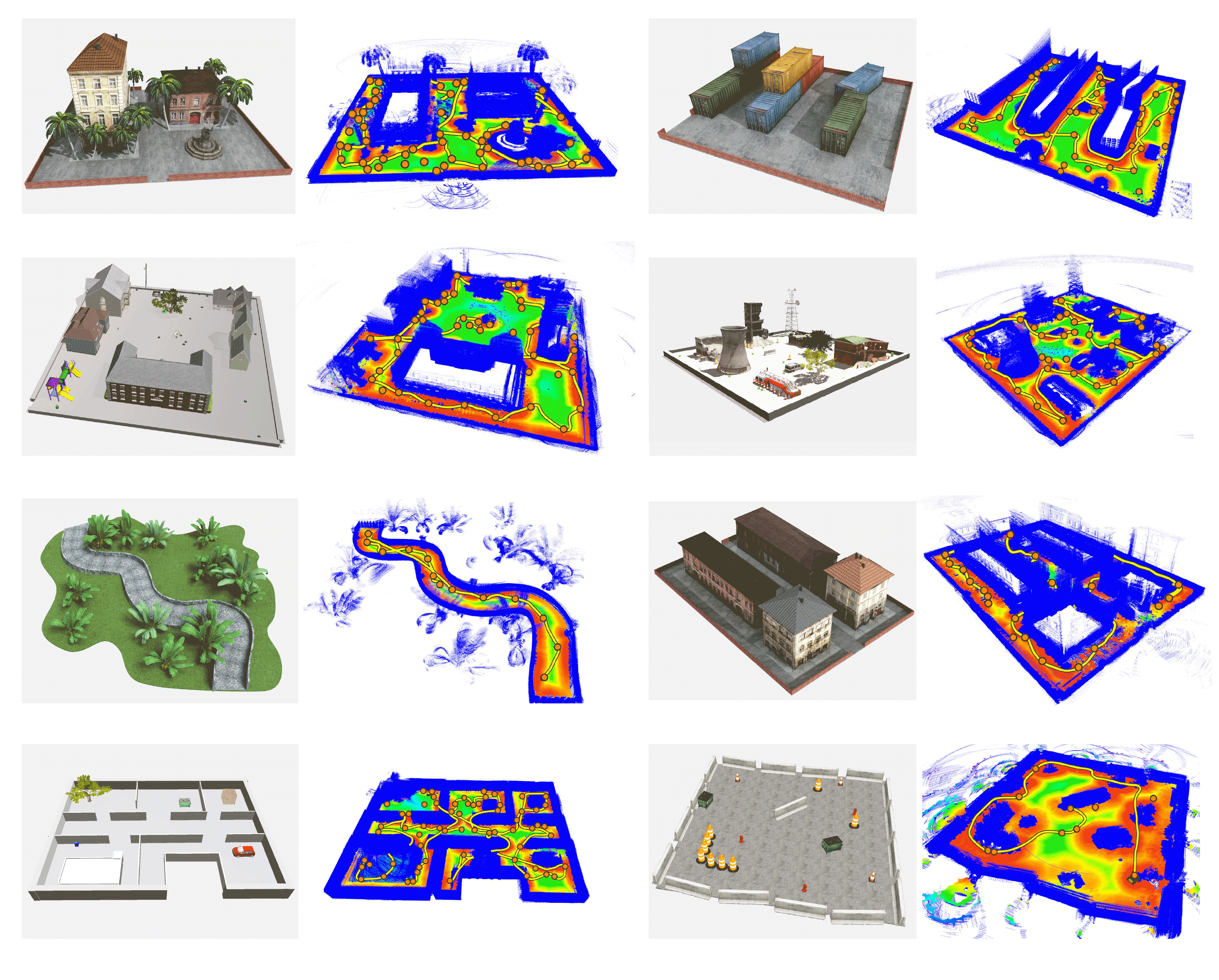}
\myfigurename{}
  	\put(2.4,60.8){\small S-scene 1}
	\put(53.4,60.8){\small S-scene 2}
	\put(2.4,40.8){\small S-scene 3}
	\put(53.4,40.8){\small S-scene 4}
	\put(2.4,20.8){\small S-scene 5}
	\put(53.4,20.8){\small S-scene 6}
	\put(2.4,1.3){\small S-scene 7}
	\put(53.4,1.3){\small S-scene 8}
\end{overpic}
\caption{
A gallery of scanned point clouds and online planned scanning paths over eight large synthetic outdoor scenes.
}
\label{fig:gallery}
\end{figure*}

\begin{figure*}[t]
\centering
\begin{overpic}
[width=\textwidth]
{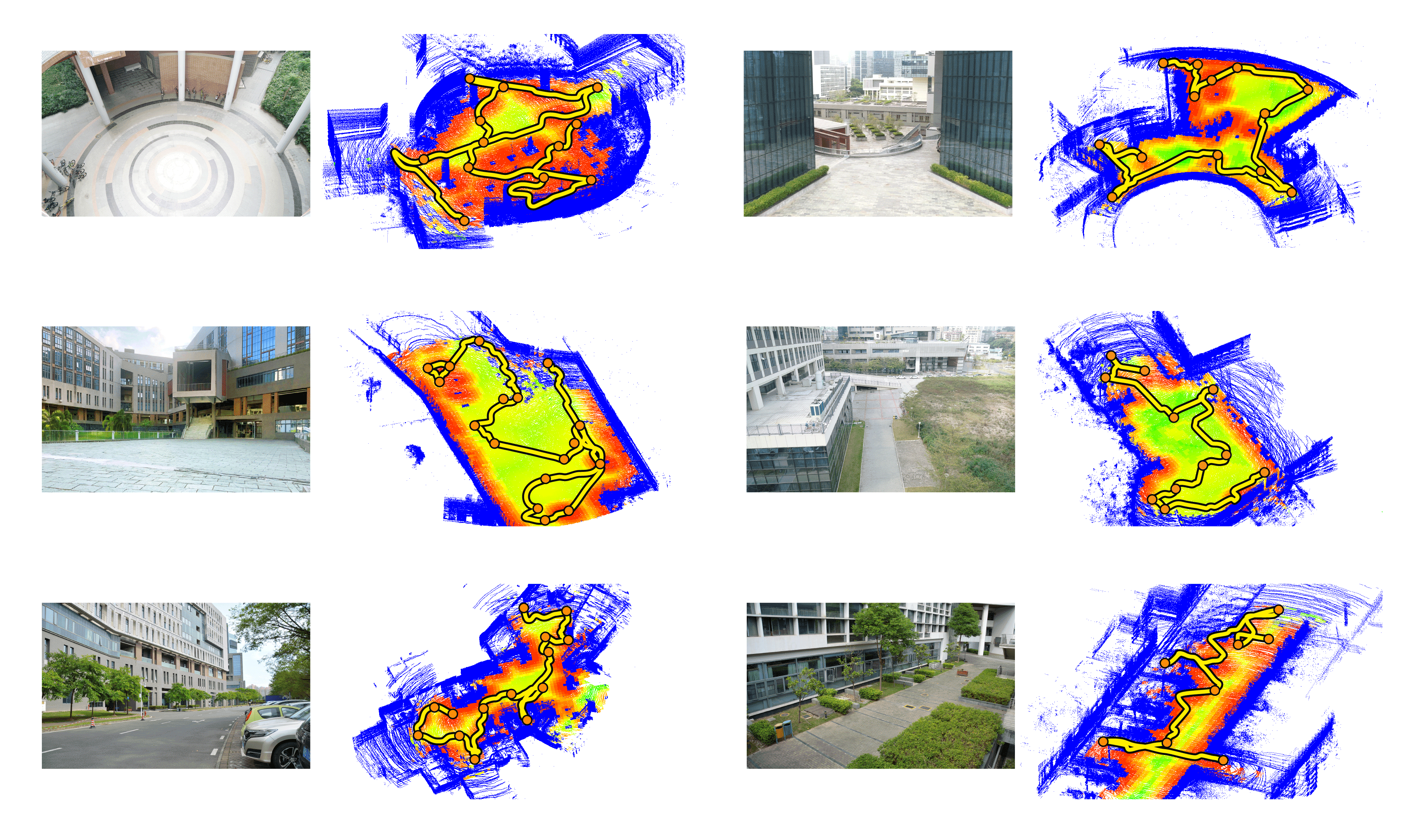}
\myfigurename{}
  	\put(2.6,42){\small R-scene 1}
	\put(53.4,42){\small R-scene 2}
	\put(2.6,22.4){\small R-scene 3}
	\put(53.4,22.4){\small R-scene 4}
	\put(2.6,2.7){\small R-scene 5}
	\put(53.4,2.7){\small R-scene 6}
\end{overpic}
\caption{
A gallery of scanned point clouds and online planned scanning paths over six real-world outdoor scenes.
}
\label{fig:realscenegallery}
\end{figure*}
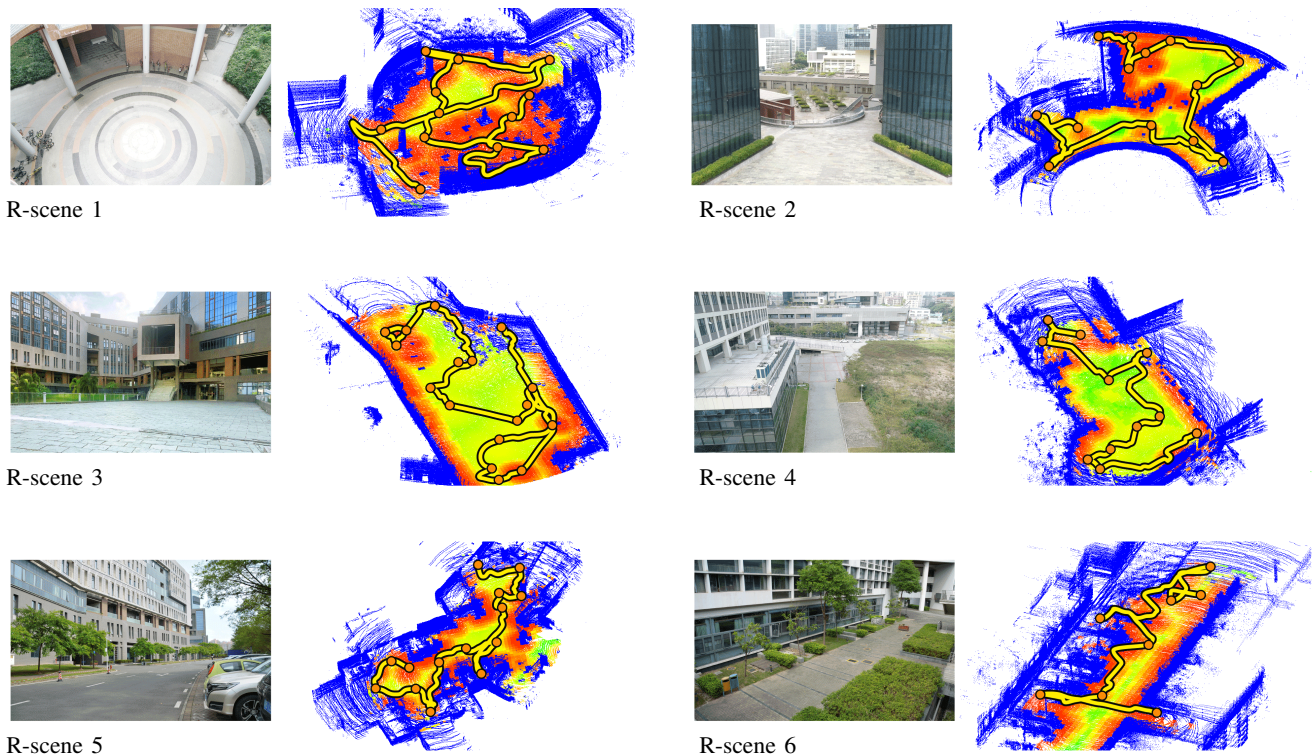

\begin{table*}
	\centering
	\caption{Average and maximum computing time (in second) of the various algorithmic components (distance-based explorability, medial-based explorability, point-wise observability, path optimization, fractal dimension computation for point cloud quality assessment, and path refinement) for five synthetic scenes. The average and maximum total computing time is reported in the last column.}
	\label{tab:time}
	\scalebox{1.18}{\setlength{\tabcolsep}{1mm}{
			\begin{tabular}{l||c|c|c|c|c|c|c|c|c|c|c|c|c|c} \hline
				\multirow{2}{*}{Scene}   & \multicolumn{2}{c|}{Distance-based trav.}   & \multicolumn{2}{c|}{Medial-based trav.}   & \multicolumn{2}{c|}{Point-wise observ.}   & \multicolumn{2}{c|}{Path optimi.} & \multicolumn{2}{c|}{Fractal dimens.} & \multicolumn{2}{c|}{Path refine.} & \multicolumn{2}{c}{Total}\\
				\cline{2-15}
				\multicolumn{1}{r||}{}         &   Avg. & Max.    &   Avg. & Max.    &    Avg. & Max.    &   Avg. & Max.   &   Avg. & Max.   &   Avg. & Max.   &   Avg. & Max.  \\ \hline\hline
				S-scene 1      &   0.12 & 0.28    &   0.51 & 1.14    &    64.3 & 145.6   &   0.12 & 0.27   &   0.42 & 0.88   &   0.22 & 0.54   &   65.7	& 148.8
				\\ \hline      S-scene 2      &   0.17 & 0.19    &   0.73 & 0.76    &    85.4 & 94.0    &   0.17 & 0.19   &   0.66 & 0.72   &   0.38 & 0.41   &   87.5	& 96.3
				\\ \hline      S-scene 3      &   0.04 & 0.16    &   0.16 & 0.21    &    36.4 & 51.0    &   0.05 & 0.07   &   0.17 & 0.22   &   0.10 & 0.13   &   36.9 & 51.7
				\\ \hline      S-scene 4      &   0.19 & 0.23    &   0.83 & 1.01    &    94.5 & 130.4   &   0.21 & 0.27   &   0.60 & 0.74   &   0.40 & 0.51   &   96.7	& 133.1
				\\ \hline      S-scene 5      &   0.16 & 0.18    &   0.68 & 0.78    &    90.8 & 110.8   &   0.15 & 0.17   &   0.57 & 0.67   &   0.17 & 0.20   &   92.5 & 112.8
				\\ \hline
			\end{tabular}
	}}
\end{table*}

\begin{figure}[t]
	\centering
	\begin{overpic}
		[width=1.0\linewidth]{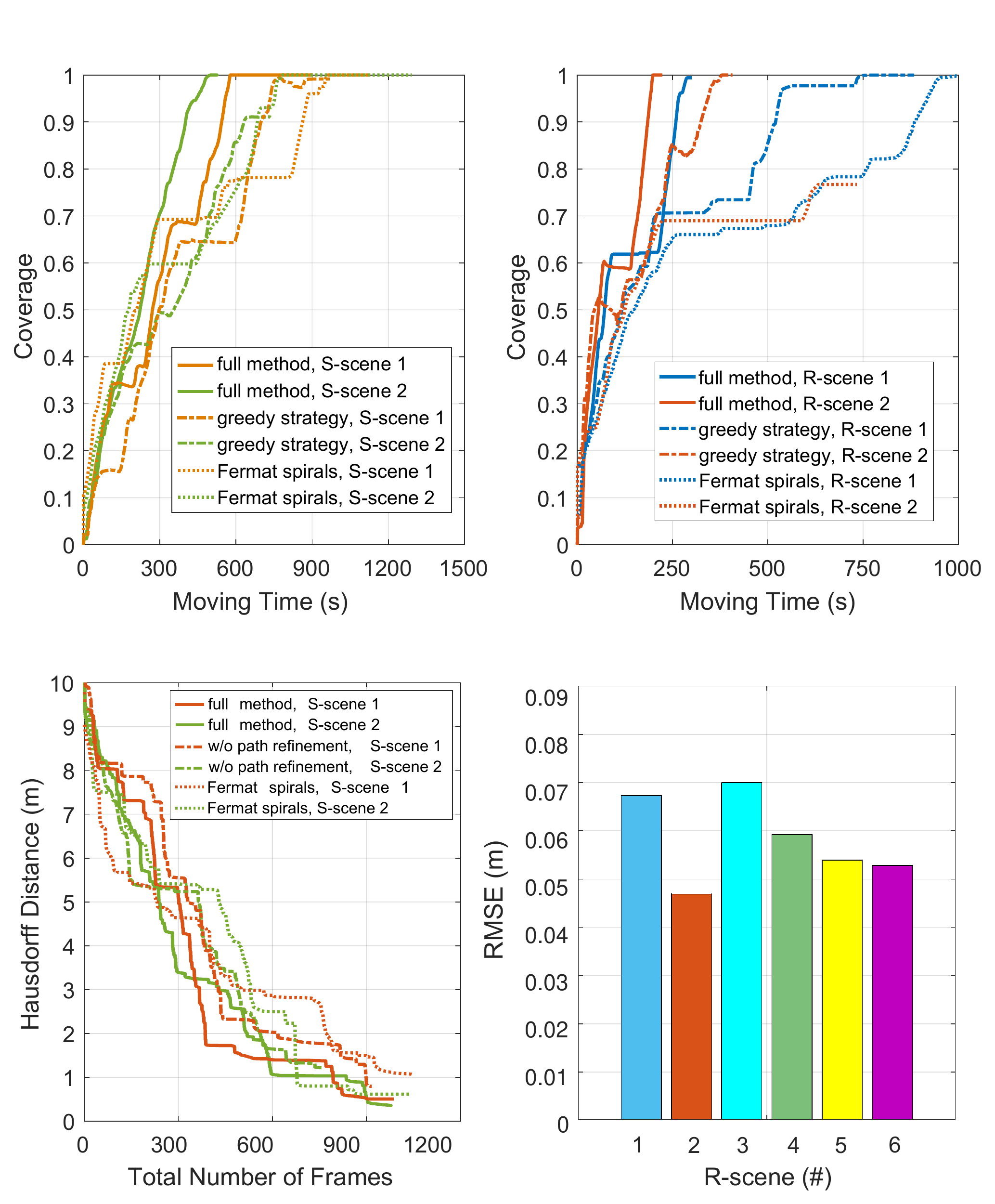}
		\put(14,95){\small (a) Synthetic}
		\put(56,95){\small (b) Real-world}
		\put(14,44.5){\small (c) Synthetic}
		\put(56,44.5){\small (d) Real-world}
	\end{overpic}
	\caption{
		Plots of coverage rate (a-b) and scanning accuracy (c-d) on both synthetic and real scenes over increasing scanning time.
	}
	\label{fig:quality}
\end{figure}

\begin{figure}
	\centering
	\begin{overpic}
		[width=0.99\linewidth]{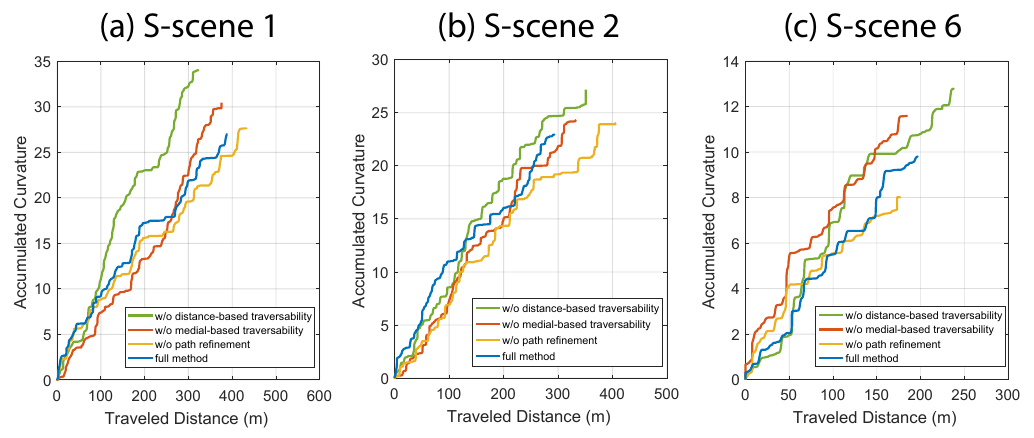}
	\end{overpic}
	\caption{
		Plots of accumulated curvature of planned paths over increasing travel distance
		on S-scene 1, S-scene 2 and  S-scene 6, respectively.
	}
	\label{fig:curvature}\vspace{-10pt}
\end{figure}

\begin{figure}[t]
\centering
\begin{overpic}
[width=0.99\linewidth]{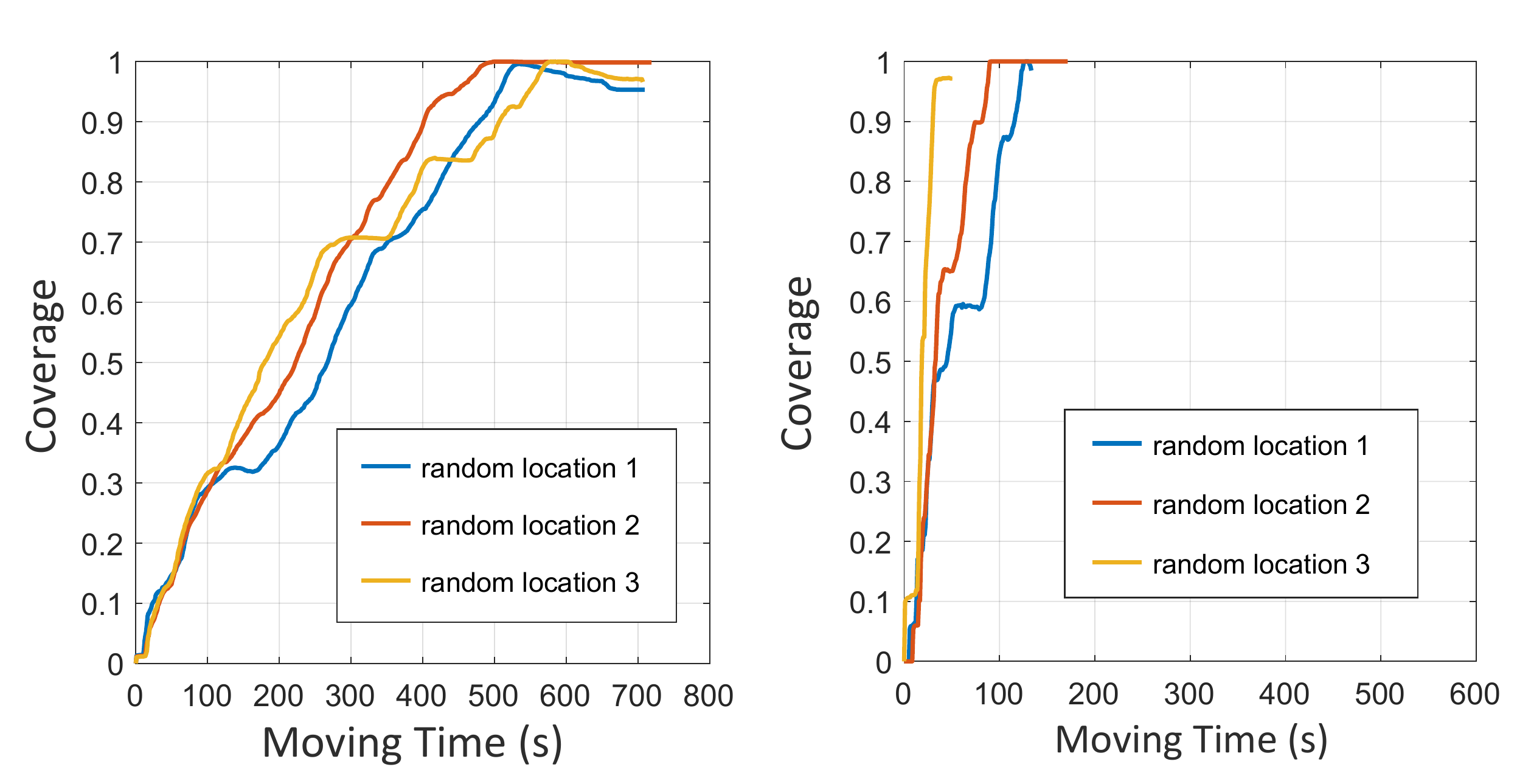}
\put(16,49){\small (a) S-scene 1}
\put(68.5,49){\small (b) S-scene 8}
\end{overpic}
\caption{
Plots of coverage rate over scanning time, for three random initializations of robot position, tested over S-scene 1 and S-scene 8.
}
\label{fig:randinit}
\end{figure}

\begin{figure}
\centering
\begin{overpic}
[width=0.49\linewidth]{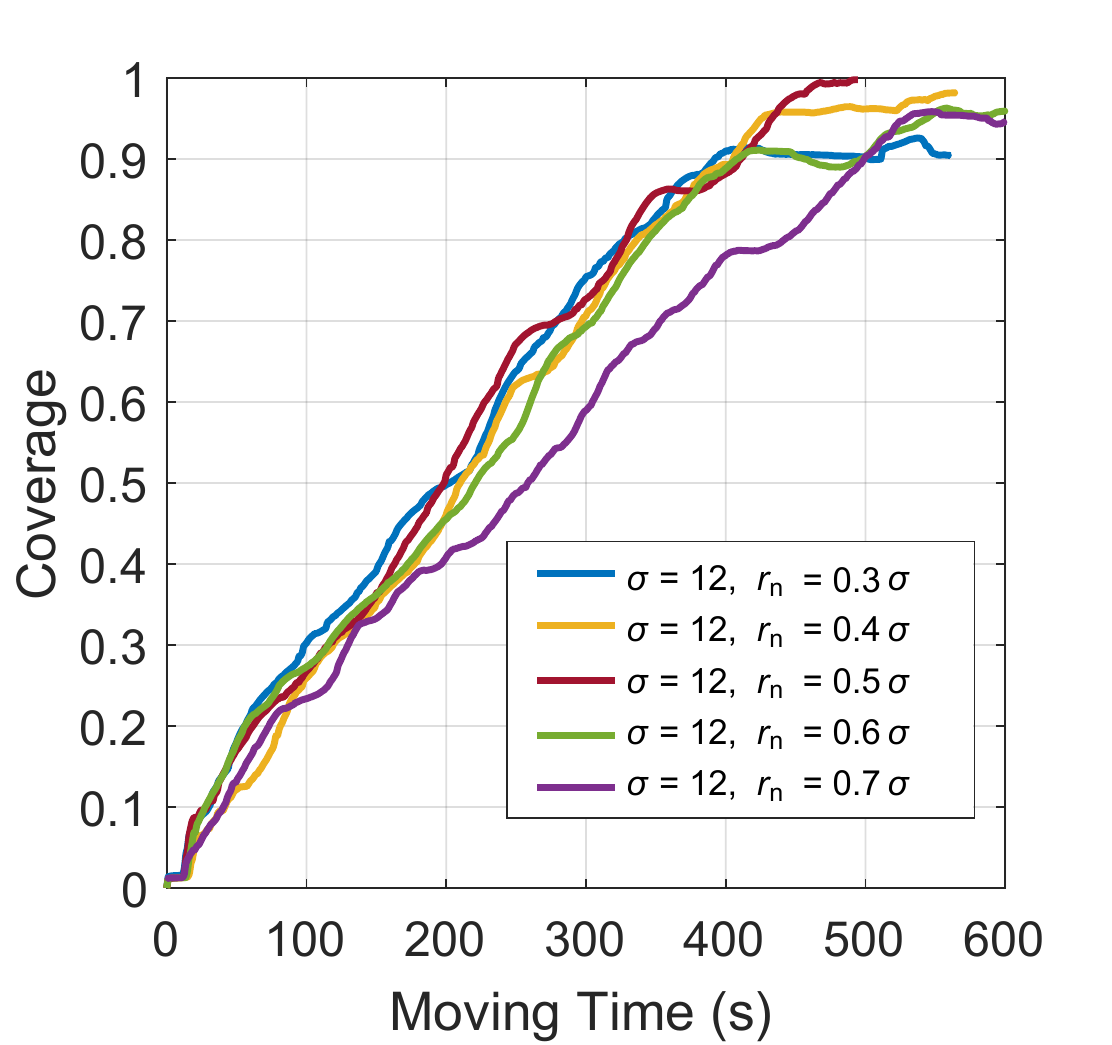}
\put(23,90){\small (a) Fixing $\sigma$ to $12$}
\end{overpic}
\begin{overpic}
[width=0.49\linewidth]{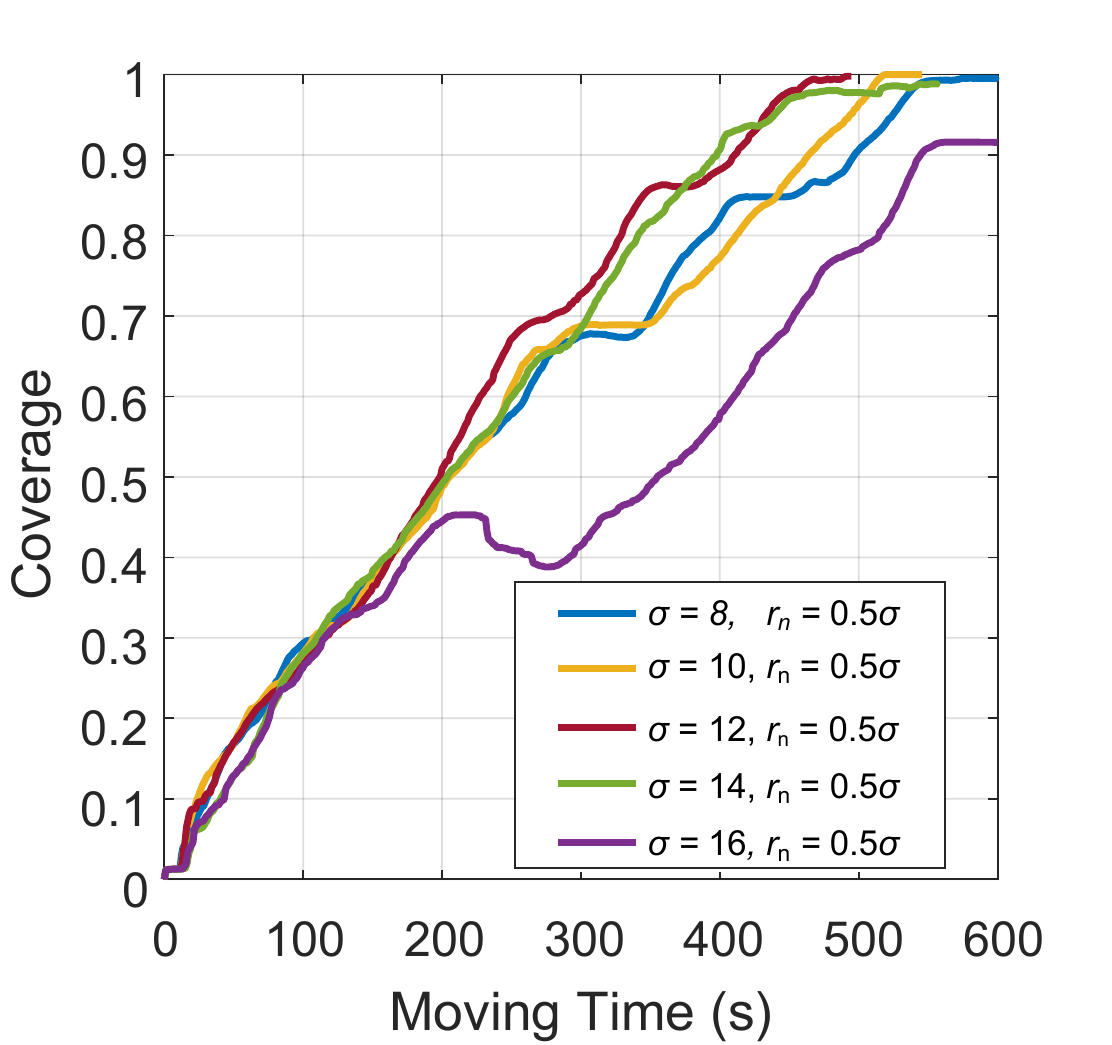}
\put(19,90){\small (b) Fixing $r_n$ to $0.5\sigma$}
\end{overpic}
\caption{
Study on the effect of two key parameters $\sigma$ and $r_n$ on coverage rate over increasing moving time, tested on S-scene 6.
}
\label{fig:param}
\end{figure}

\begin{figure}[b]
	\centering
	\begin{overpic}
		[width=0.49\linewidth]{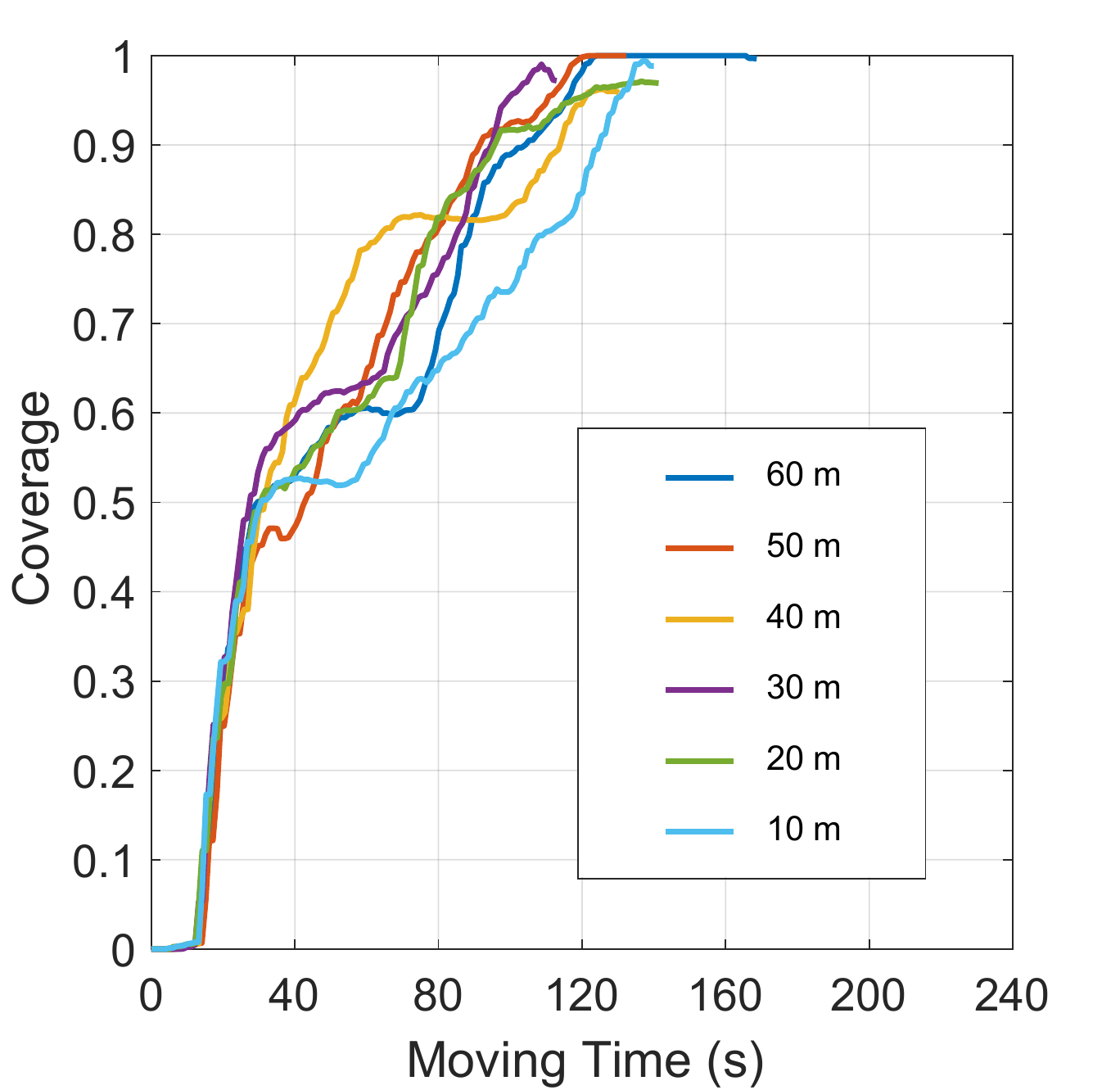}
		\put(32,95){\small (a) S-scene 8}
	\end{overpic}
	\begin{overpic}
		[width=0.49\linewidth]{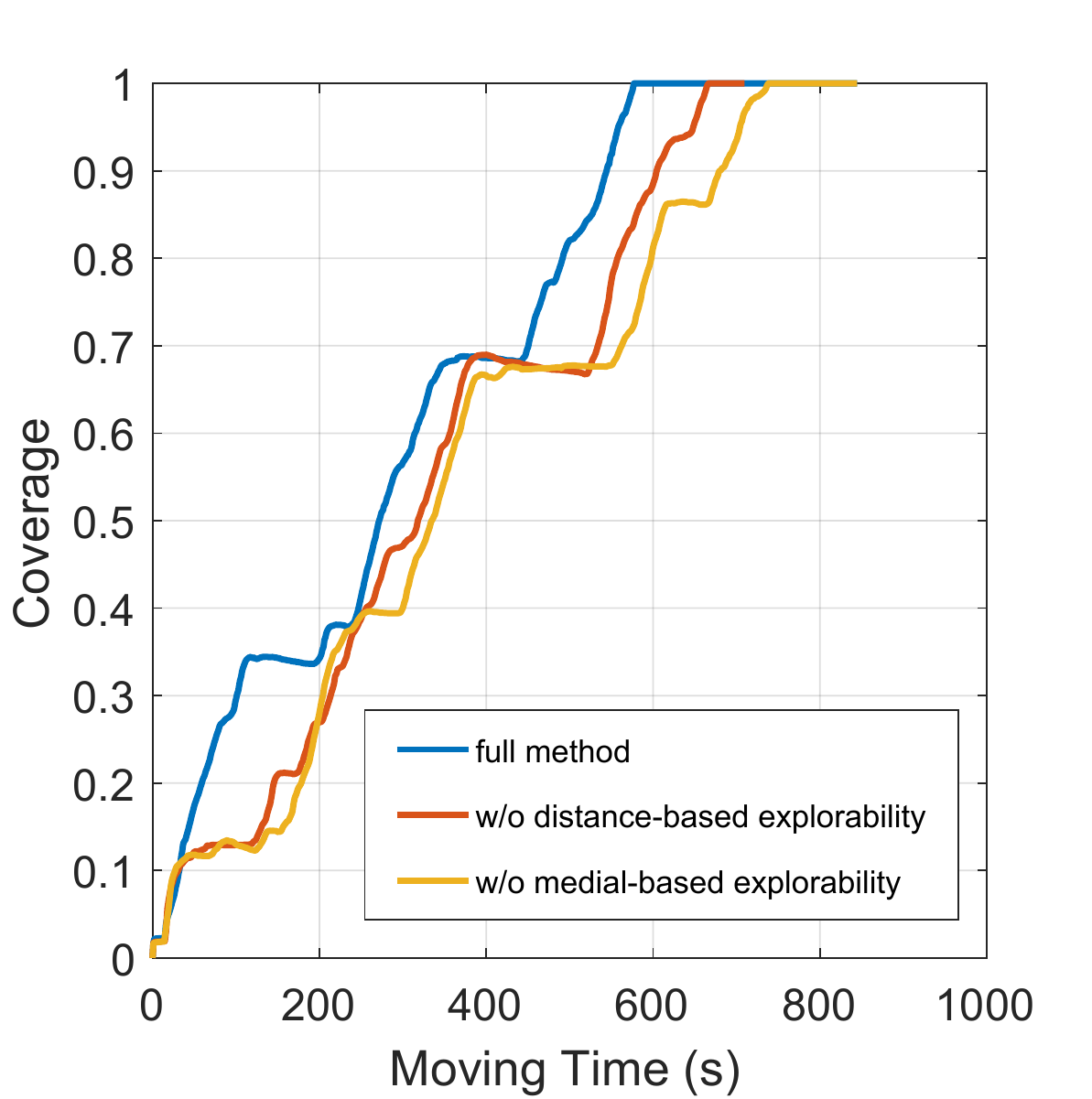}
		\put(30,95){\small (b) S-scene 1}
	\end{overpic}
	\caption{
		(a): A study of scanning efficiency with different viewing distance of LiDAR sensor, tested on S-scene 8. (b): An ablation study of our method without the distance-based and medial-based explorability terms in the computation of topological guidance map, tested on S-scene 1.
	}
	\label{fig:ablation_viewdistance}
\end{figure}

\paragraph{Dataset}
Our method has been carefully evaluated both on synthetic and real-world scenes, as shown in \Fig{gallery} and \Fig{realscenegallery}, respectively.
The synthetic dataset contains $8$ scenes, among which $4$ are composed of narrow passages,
$4$ are squares with obstacles.
For synthetic scanning, we adopt the virtual laser scanning provided by Velodyne Lidar, Inc., which accounts for scanning noise and imperfect reproducibility.

We also perform field test on six real-world scenes.
For each scene, we let a human operator perform an as-complete-as-possible scanning for a prescribed region of an outdoor environment. For each scanned point cloud, we extract the ground serving as the ground-truth for measuring scanning completeness and quality.
\Table{stats} reports the statistics of the six real-world outdoor scenes used for field test. Note that most of the real-world scenes are unbounded open areas.
For each scene, we report the total area our robot scanned and the total traverse distance and time for finishing the scanning.

\paragraph{Metrics}
We carefully evaluate three aspects of our autonomous scanning: 1) \emph{Coverage completeness} computed as the Intersection-over-Union (IoU) between the scanned point cloud and the corresponding ground-truth; 2) \emph{Scanning accuracy} measured by Hausdorff distance for synthetic data and by Root Mean Squared Error (RMSE)
for real-world data; 3) \emph{Path smoothness} measured by integrated curvature along the robot path.

\paragraph{Timing information}
\Table{time} reports the average and maximum computing time (in second) of the various algorithmic components for five synthetic scenes. The physical moving time of our Husky robot is not included.
\Table{stats} (last column) also reports the total scanning time for the six real-world scenes.

\paragraph{Coverage completeness}
At the top of \Fig{quality}, we plot the completeness of scanning coverage over increasing robot moving time
for our method and two alternatives (greedy and Fermat spiral~\cite{wu2019energy}).
The plots show that our method achieves a faster coverage for both synthetic (\Fig{quality}(a)) and real-world (\Fig{quality}(b)) scenes.
Note the coverage may sometimes decrease because some points that are found
to be very close to a newly discovered obstacle are removed from the already scanned point cloud, for the sake of collision avoidance. In the robustness against random initialization part below, we show more evaluations of scanning coverage with different initializations of robot location.

\begin{figure*}
\centering
\begin{overpic}
[width=0.99\linewidth]
{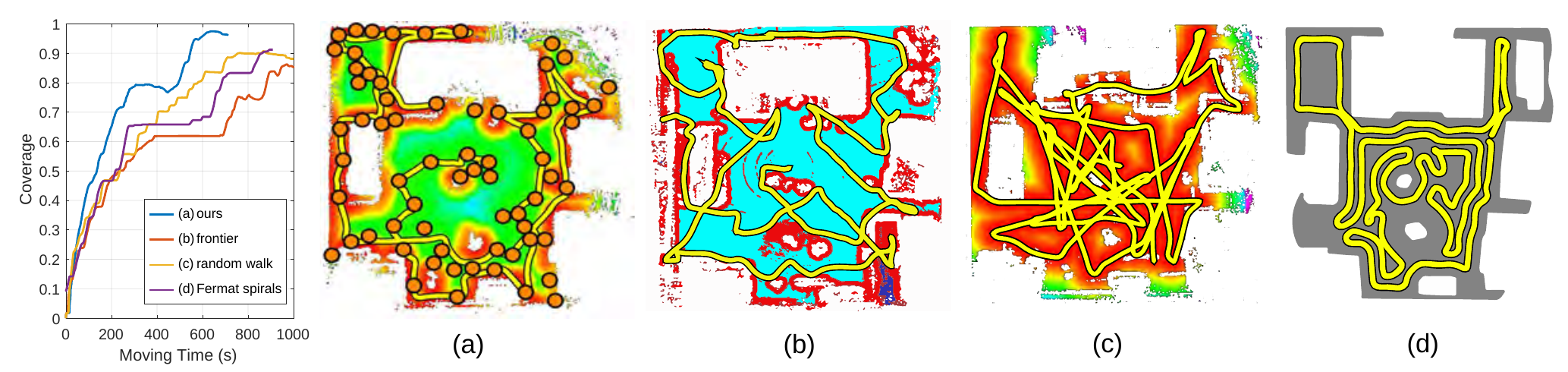}
\myfigurename{}
\end{overpic}
\caption{
Comparing the coverage rate in auto-scanning S-scene 3 by our method, frontier exploration, random walk and Fermat spiral coverage. The scanned point clouds and traverse paths for each method are shown in right.
}
\label{fig:comparison}
\end{figure*}

\begin{figure*}
\centering
\begin{overpic}
[width=0.99\textwidth]
{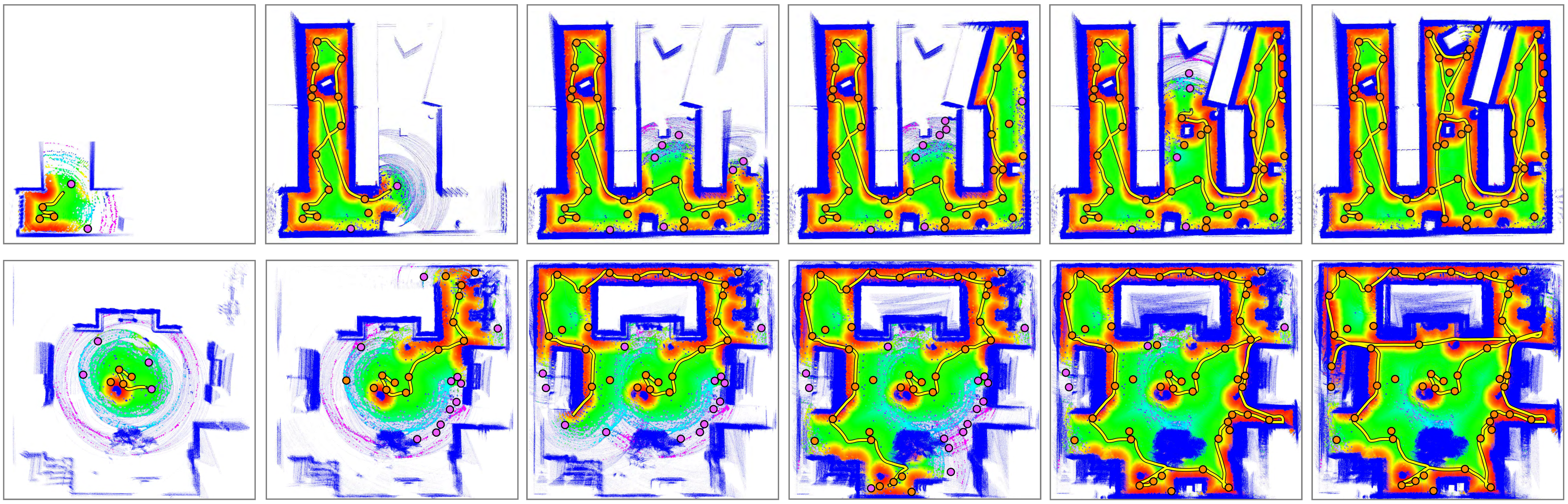}
\myfigurename{}

\end{overpic}
\caption{
Demonstrating the progressive scanning of S-scene 2 (top) and S-scene 3 (bottom) by our system, with a series of snapshots of
acquired point clouds. Over each point cloud, we plot the color-coded scanning confidence, the visit sites and the planned paths.
}
\label{fig:progression}
\end{figure*}

\paragraph{Scanning accuracy}
\Fig{quality}(c) shows the scanning accuracy
over increasing scanning frames with three different synthetic scenes.
The accuracy is measured as the Hausdorff distance between the scanned point cloud and
the ground-truth mesh.
In comparison, we also plot the results of Fermat spiral as well as
an ablated version of our method, i.e., w/o path refinement.
Our robot system achieves consistently higher accuracy throughout the scanning,
demonstrating its robustness against noise, occlusion and moving objects.
\Fig{quality}(d) provides the scanning accuracy measured by RMSE on six real-world scenes.
The low RMSE values indicate that our path planning does not bring much burden on SLAM mapping while ensuring the higher coverage rate and scanning quality.

\paragraph{Path smoothness}
\Fig{curvature} shows the curvature accumulates slowly with increasing travel distance, over
three synthetic scenes.
We study the effect of the two explorability terms and the path refinement step on path smoothness.
Distance-based explorability has more influence on path quality,
because the time step is long without this term, leading to less smooth traverse paths. The path refinement step that enhances the scanning accuracy (see \Fig{quality}(c)) has some bearable impact on the final path smoothness. 

\paragraph{Robustness against random initialization}
\label{sec:rand}
\Fig{randinit} evaluates the completeness over increasing traveling time,
for different random initializations of robot location in two synthetic scenes.
The plots show that different initial locations would lead to almost the same speed of coverage,
demonstrating the insensitivity of our method to initialization.
For both scenes, our method arrives at a full coverage within $500$m and $150$m traverse distance (the traversing speed is $1$m/s) for all initial locations, respectively. Inspecting the initial random locations, the location 2 and location 3 in S-scene 8 are around the center of that scene, so the completeness raises fast initially; see the right plot of \Fig{randinit}.

\paragraph{Study on key parameters}
We study the effect of two key parameters in our method:
the radius of circular planning area $\sigma$ (\Eq{traverse-d}) and
the radius of the neighborhood used for NMS-based site selection $r_\text{n}$ (\Eq{visitsite}).
\Fig{param} plots the effect of these parameters on coverage rate.
From the results, a smaller $\sigma$ leads to more detailed scanning due to the
smaller planning region. Meanwhile, too small $\sigma$ causes slow increase of scanning coverage
due to too local planning.
Similarly, using a small $r_\text{n}$ benefits the coverage of small scale geometry, with the cost of
slow scanning. On the other hand, when a large $r_\text{n}$ is used, the robot might miss some
important topological features in the scene such as small branches.
We found that $\sigma=12$m and $r_\text{n}=6$m generally result in a good
trade-off between quality and efficiency.
Besides, we have carried out an efficiency study depend on sampling the viewing distance, i.e., the farthest scanning distance of LiDAR in \Fig{ablation_viewdistance}(a). The result shows that the viewing distance has no much effect on the algorithm efficiency, as the proposed method only considers the points within given valid range, which is loose for mobile robots and LiDAR sensors.

\paragraph{Ablation study}
To verify the importance of the distance-based explorability and medial-based explorability
terms in the computation of topological guidance map,
we conduct an ablation study via disabling each of them and observe the effect on coverage rate.
From the plots in \Fig{ablation_viewdistance}(b), when distance-based explorability is disabled,
the coverage rate increases slower with a lower convergence value.
This is because the robot would keep away from boundaries (walls) and can miss some branches
without the distance-based explorability term.
When medial-based explorability is removed, the robot tends to move towards boundaries
rather than open areas. Also please note that S-scene 1 shown in \Fig{gallery} is $2500$m$^2$ while the travelable open area inside is about $1840$m$^2$, including width streets and a fountain square. The three buildings that are as the main obstacles in this scene are indeed sparse, and thus the robot path may likely exhibit some oscillations, resulting in slow coverage increase as what can be observed in \Fig{ablation_viewdistance}(b).
Therefore, both explorability terms are essential in reasonably planning the scanning course.

\paragraph{Comparison}
We are not aware of a practical solution for outdoor scanning with
the goal of scene coverage and scanning quality.
Therefore, we compare our method with random walk, frontier based exploration, and the most recent Fermat spiral based coverage~\cite{wu2019energy}.
In random walk, the robot chooses a random move at each time step.
The frontier-based scanning takes the standard implementation provided in ROS, where frontiers are extracted
based on the OctoMap~\cite{Hornung13}.
The plots in \Fig{comparison} show that our method achieves the fastest increase of coverage, based on the online topological and geometrical path planning.
For outdoor scanning, frontier-based method does not perform as well as for indoor scanning, because
the range of frontier could spread immensely for open area making it less useful in guiding robot exploration. In addition, too many frontier points would lead to back-and-forth traverse, which degenerates to random walk. Fermat spirals method requires that the scene is known, so not autonomous. It aims for full path coverage of the provided scene, which very likely leads to redundant traverse with respect to remote LiDAR scanning; see also the accompanying video.

\paragraph{Qualitative results}
\Fig{gallery} and \Fig{realscenegallery} show two galleries of scanned point clouds and online planned scanning paths for eight synthetic and six real-world outdoor scenes, respectively. The non-ground points (e.g., walls and other obstacles) are shaded in blue.
Over the ground points, we plot the final \emph{confidence field} which is inversely related to the point-based
guidance field, i.e., $1-\tau(p)$. Red indicates to high confidence.
The purple dots represent the visit sites selected and visited by the robot.
The traverse path connecting the sites is visualized as a yellow curve.
Note that the scene boundary (such as walls if any) is unknown to the robot for all examples.

In \Fig{progression}, we show two examples of the progressive scanning process by our system.
We demonstrate a series of snapshots of progressively acquired point clouds, the dynamic changing visit sites
and the online planned traverse paths.
Note how our method is able to dynamically determine a proper set of visit sites and plan a cost-efficient scanning path through them.
See also live demonstration of our system in real-world scanning in the accompanying video.

\section{Conclusion}
\label{sec:future}
We have presented a robust algorithm for on-the-fly planning of robot scanning, based on
a discrete-continuous optimization of exploration paths. The optimization interleaves between
topological guidance map construction, online path finding, and geometrical refinement of traverse path.
Based on the algorithm, we implement a robust system for autonomous scanning of outdoor environments.
The method attains the following key features.
First, it conducts path planning based on the point cloud within the planning area around the robot, without the need of a volumetric occupancy map which is computationally intractable for outdoor scenes.
Second, it achieves unknown region estimation based on a key assumption of visibility continuity, without an occupancy map.
Last but not least, it realizes quality-driven path planning with the robust fractal dimension based quality measurement.

\paragraph{Limitations and future work}
Our method and the system have the following limitations, over which we would like to discuss potential improvements for future work:
\begin{itemize}
	\item \emph{Prediction mechanism}. Our planning is conducted over the point cloud around the robot, which is local in nature. To gain more global planning, one could consider predicting the explorability and observability for a larger range of point cloud, based on the scan points database, using learning-based methods.
	\item \emph{Handling dynamic scenes}. Our current method does not handle dynamic scenes with moving objects, such as cars and pedestrians, since the latter causes \emph{ghosting points} that might be mistakenly regarded as obstacles. In future, we would like to enhance our method with moving object detection and removal.
	\item \emph{Physical constraints}. Our geometric path refinement does not integrate the physical DoF constraints of the robot. In our current implementation, DoF constraints are resolved after the path optimization. Thus, physically unresolvable paths do occur, although quite seldom.
	\item \emph{Learning mechanism}. Our current development does not include a learning mechanism. We would look into deep reinforcement learning to plan for outdoor scanning. The planned paths by our method could be used to guide an efficient policy learning.
\end{itemize}

\bibliographystyle{IEEEtran}
\bibliography{husky}

\end{document}